\documentclass[twoside]{article}

\usepackage[accepted]{aistats2022}
\usepackage[round]{natbib}

\usepackage{header}

\newcommand{\ui}{\mathrm{bayes}}
\newcommand{\spl}{\mathrm{boost}}

\newcommand{\stackGP}{\textsc{MHGP}\xspace}
\newcommand{\DEFstackGP}{Mean Hierarchical GP\xspace}

\newcommand{\hierarchicalGP}{\textsc{HGP}\xspace}
\newcommand{\DEFhierarchicalGP}{Hierarchical GP\xspace}

\newcommand{\seqhierarchicalGP}{\textsc{SHGP}\xspace}
\newcommand{\DEFseqhierarchicalGP}{Sequential Hierarchical GP\xspace}

\newcommand{\multitaskGP}{\textsc{MTGP}\xspace}
\newcommand{\DEFmultitaskGP}{Multi-Task GP\xspace}

\newcommand{\multitasksinglekGP}{\textsc{MTKGP}\xspace}
\newcommand{\DEFmultitasksinglekGP}{Multi-Task-Single-$k$ GP\xspace}

\newcommand{\boostedGP}{\textsc{BHGP}\xspace}
\newcommand{\DEFboostedGP}{Boosted Hierarchical GP\xspace}

\newcommand{\weightedSourceGP}{\textsc{WSGP}\xspace}
\newcommand{\DEFweightedSourceGP}{Weighted Source GP\xspace}

\newcommand{\x}{\mathbf{x}}
\newcommand{\X}{\mathbf{X}}

\newcommand{\funcdist}[3]{f_{{#2}}^{{#1}}{#3}}

\newcommand{\uidist}[2]{\funcdist{\ui}{#1}{#2}}
\newcommand{\spldist}[2]{\funcdist{\spl}{#1}{#2}}

\newcommand{\expectation}[1]{\mathbb{E}\left(#1\right)}

\newcommand{\cov}[1]{\mathrm{cov}\left(#1\right)}
\newcommand{\var}[1]{\mathrm{var}\left(#1\right)}

\newcommand{\abs}[1]{\left|#1\right|}

\newcommand{\normalPDF}[3]{p\left(#1; #2, #3\right)}
\newcommand{\GP}[2]{\mathcal{GP}\left(#1, #2\right)}
\newcommand{\normaldist}[2]{\mathcal{N}\left(#1, #2\right)}

\newcommand{\woodburry}{\alpha_{*,t}}

\begin{document}

%

%

\twocolumn[

\aistatstitle{Transfer Learning with Gaussian Processes for Bayesian Optimization}

\aistatsauthor{Petru Tighineanu \And Kathrin Skubch \And  Paul Baireuther}
\aistatsauthor{ Attila Reiss \And Felix Berkenkamp \And Julia Vinogradska}
\aistatsaddress{Bosch Center for Artificial Intelligence, Renningen, Germany }

\runningauthor{Tighineanu, Skubch, Baireuther, Reiss, Berkenkamp, Vinogradska}

]

\doparttoc 
\faketableofcontents 

\begin{abstract}
  Bayesian optimization is a powerful paradigm to optimize black-box functions based on scarce and noisy data. Its data efficiency can be further improved by transfer learning from related tasks.
  While recent transfer models meta-learn a prior based on large amount of data, in the low-data regime methods that exploit the closed-form posterior of Gaussian processes (GPs) have an advantage.
  In this setting, several analytically tractable transfer-model posteriors have been proposed, but the relative advantages of these methods are not well understood.
  In this paper, we provide a unified view on hierarchical GP models for transfer learning, which allows us to analyze the relationship between methods. As part of the analysis, we develop a novel closed-form boosted GP transfer model that fits between existing approaches in terms of complexity.
  We evaluate the performance of the different approaches in large-scale experiments and highlight strengths and weaknesses of the different transfer-learning methods.
\end{abstract}

\section{INTRODUCTION}
Bayesian optimization (BO) is an elegant and powerful approach to black-box optimization. It has been successfully applied to several challenging black-box optimization problems, ranging from hyperparameter optimization~\citep{snoek_hpo} to materials design \citep{zhang2020} and controller tuning \citep{calandra2016bayesian}. 
One major advantage of BO is sample efficiency: it is specifically tailored to expensive black-box functions, where each function evaluation is costly, e.g., when laborious experiments on physical systems are involved. For such applications, the benefit of fewer function evaluations outweighs the increase in computational cost to make informed decisions about where to measure next.
A key ingredient to BO's sample efficiency is a probabilistic surrogate model of the objective. In the low-data regime, the most commonly employed model is a Gaussian process (GP) \citep{rasmussengp}, as it provides closed-form posteriors with accurate uncertainty estimates to guide the search for the global optimum.

Many black-box optimization problems are not one-off tasks, but rather several related instances of the tasks can be encountered. Here, the data efficiency of optimization can be further improved by transferring knowledge from related tasks. Especially in the low-data regime that BO is specialized on, transfer learning provides great value. To this end, \citet{NIPS2013_swersky,EnvGP,pmlr-v54-shilton17a,golovin2017stackgp} propose several approaches to transfer learning for BO. These transfer learning models combine related task data and the data of the current task into a joint model that guides BO in the search for the current task's global optimum. This setting implies a certain asymmetry: while we may use data from related tasks, we are only interested in informative models for the current task. Further, the related task data are considered as given and additional samples cannot be acquired. One common approach to account for this asymmetry, often referred to as \emph{hierarchical model}, is to model the difference between the current and related tasks. However, when combining different task data into a single GP model, the computational complexity scales cubically with the number of data points. This complexity can easily become prohibitive if data from multiple related tasks are available. Several other transfer models with tractable posteriors and more favourable scaling have been proposed, e.g., an ensemble of GPs~\citep{feurer2018rgpe} or a hierarchical model for the mean prior~\citep{golovin2017stackgp}. However, their relation to the joint models with full cubic complexity, their relative strengths and weaknesses, and underlying assumptions are not well understood.

\section{RELATED WORK AND CONTRIBUTIONS}
The problem of speeding up Bayesian optimization by using additional information besides the objective function evaluations is a focus topic in the black-box optimization community. A large body of literature promotes the idea to meta-learn models \citep{pmlr-v70-finn17a,ABLR,flennerhag2018transferring}, whole optimizers \citep{pmlr-v70-chen17e} or acquisition functions \citep{Volpp2020Meta-Learning} for BO. These methods are powerful, but require large amounts of training data to extract the characteristics of a class of tasks. 

In the low-data regime, methods that rely on the uncertainty estimates of a GP posterior to guide the search outperform large-scale meta-learning. Several GP-based transfer learning approaches have been proposed, which can mostly be summarized into two categories that either \textit{(i)} build a global GP model, or \textit{(ii)} maintain separate GPs for each data set. Several options exist to aggregate this data into a single global model: multitask GPs by \citet{NIPS2013_swersky,pmlr-v33-yogatama14} model correlations between tasks, \citet{NIPS2017_MISO} model task biases with a joint linear coregionalization kernel and some independence assumptions, \citet{EnvGP} propose envelope GPs that adapt the noise model to accommodate all data, and DiffGPs by \citet{pmlr-v54-shilton17a} regress on prediction bias corrected related task data.

In multi-fidelity BO, which only differs from the transfer learning setting by removing the constraint that related task data cannot be acquired during optimization, \citet{forrester2007multi,Marco17VirtualvsReal} successfully employ GP models. All these approaches ultimately compute a GP model on a data set of the size of the joint related and target task data sets and suffer from the cubic complexity. On the other hand, \citet{golovin2017stackgp}, \citet{feurer2018rgpe} and \citet{Wistuba} maintain separate GPs for each data set to avoid the accumulated scaling. In the hierarchical model presented in \citet{golovin2017stackgp}, this reduction in computational complexity stems from neglecting the uncertainty of the source models for the transfer, which is detrimental to the optimization process. Instead, RGPE~\citep{feurer2018rgpe} builds a ranking-based ensemble of the separate GPs, which involves careful tuning of the weight estimator and also does not result in a Bayesian treatment of the uncertainty. In light of the described properties of these two classes of GP-based transfer learning models, the need for models that can bridge the gap between fully considering source uncertainty and not considering it at all becomes evident.
Such an intermediate regime may be particularly relevant for optimizing monetarily expensive processes with relatively little duration such as destructive measurements.
In the engineering community, efficient hierarchical models have been developed for the special case where related tasks are evaluated on decreasing subsets of inputs, see e.g. \cite{kennedy2000predicting, le2014recursive}.
However, this setting is not suitable for transfer learning from generic historical data.

While computationally efficient approximations to GPs exist that could be applied to any GP model (including those proposed in this paper), e.g. by \citet{lazaro2010sparse}, this is an orthogonal research direction that is agnostic to the structure of the underlying data and treats every data point the same. Contextual optimization \citep{krause2011contextual} is another related line of research in which context variables influence the process to-be-optimized. By contrast, transfer learning leverages discrete historical tasks as context.

\paragraph{Contributions}
In this paper, we provide a unifying view on GP based transfer learning models and shed light on their relation and assumptions. This unified view enables a thorough analysis of the methods: their computational complexity, their modelling assumptions and their respective advantages for different transfer scenarios. In addition, we fill the gaps in this unified framework. 
Firstly, we present a modular version of \DEFhierarchicalGP that we name \emph{\DEFseqhierarchicalGP} with improved scaling while maintaining competitive performance. Secondly, we develop a new transfer learning model that is a middle ground between existing approaches in terms of computational complexity and optimization performance. This method is conceptually inspired by boosting architectures, and yet results in an analytically tractable posterior, which we name \emph{\DEFboostedGP}. A comprehensive empirical evaluation of all methods supports our analytical insights and provides valuable guidance on the different methods' comparative advantages.

\section{PROBLEM STATEMENT}
\label{sec:problem_statement}

Our goal is to find the optimum of a \emph{target} black-box function $f_t \colon D \to \mathbb{R}$, based on noisy observations $\mathcal{D}_t = \{ \x_n, y_n \}_{n=1}^{N_t}$ where each observation $y_n = f_t(\x_n) + \varepsilon_n$ is corrupted by \iid zero-mean Gaussian noise, $\varepsilon_n \sim \mathcal{N}(0, \sigma_t^2)$. We model our prior belief over the target function $f_t$ with a GP \citep{rasmussengp},
$
 f_t \sim \GP{m}{k},
$
with mean function $m(\cdot)$ and kernel function $k(\cdot, \cdot)$. Conditioned on the data, the posterior distribution is a GP again with the posterior mean and variance at a query point $\x_*$
\begin{equation*}
\begin{split}
\label{eq:basic_posteriors}
\expectation{f_t \mid \x_*, \mathcal D_t}
    &=m(\x_*) + k(\x_*, \X_t)\\
    &\times\left(k(\X_t, \X_t) + \sigma_t^2 \mathbf{I}\right)^{-1} \left(\mathbf{y}_t-m(\X_t)\right), \\
\var{f_t \mid \x_*, \mathcal D_t}
	&=  k(\x_*, \x_*) - k(\x_*, \X_t )\\
	&\times\left(k(\X_t, \X_t) +\sigma_t^2\mathbf{I}\right)^{-1} k(\X_t, \x_*),
\end{split}
\end{equation*}
where $\X_t = (\x_1, \dots, \x_{N_t})$ and $\mathbf{y}_t = (y_1, \dots, y_{N_t})$ is the vector of corresponding, noisy observations. Given this belief over the function $f_t$, BO aims to actively select a new query point $\x_{N_t + 1}$ that is informative about the optimum of $f_t$. This requires trading off exploitation and exploration and is usually accomplished through an acquisition function $\alpha$ that depends on the posterior, 
\begin{equation}
    \x_{N_t + 1} = \argmax_{\x \in D} \alpha( f_t \mid \x, \mathcal{D}_t ).
\end{equation}
Common choices for $\alpha$ are the expected improvement \citep{jones1998efficient} and the upper confidence bound \citep{srinivas10gaussian}, among several others. The data-efficiency of these methods crucially depends on how fast the posterior distribution collapses around the true target function $f_t$. 

To enable faster optimization, transfer learning additionally exploits data from $n_s \geq 1$ related \emph{source} tasks based on functions $f_s\colon D \rightarrow \mathbb{R}$. While all methods discussed in the following apply to this general setting, we focus on $n_s = 1$ for ease of exposition.
In this case, we have $N_s$ additional data points $\mathcal D_s= \{\X_s, \mathbf{y}_s\}$ that can be used to improve the target model. 
Further, we assume without loss of generality that the source is modelled with a GP, $f_s \sim \GP{0}{k_s}$, with zero mean prior and arbitrary kernel function $k_s.$

\section{GAUSSIAN PROCESS TRANSFER MODELS}
\label{sec:models}

In this section, we provide a unified overview of existing GP models for transfer learning that allow for a closed-form posterior distribution. 

\paragraph{\DEFmultitaskGP (\multitaskGP)} Our starting point is the kernel function $k$ of the \emph{joint} model of source and target. 
Let $(\x, i), (\x', j)$ be two points from tasks $i,j \in \{ s, t \}$. We assume that $k$ is a sum of separable kernels
\begin{equation}
\label{eq:general_kernel}
    k(( \x, i), ( \x', j)) = \sum_{\nu\in \{ s, t \}} [\mathbf{W}_\nu]_{i, j} \, k_\nu( \x, \x') + 
    \delta_{\x \x'}\delta_{ij}\sigma_i^2 ,
\end{equation}
where the dirac-delta function $\delta_{ij}$ is equal to one if $i=j$ and zero otherwise, and $k_\nu$ are arbitrary kernel functions \citep{alvarez2012kernels}. The positive semi-definite matrices $\mathbf{W}_\nu$ are often referred to as \emph{coregionalization matrices}, since their diagonal (off-diagonal) entries control correlations within (between) data sets. 
When counting the hyperparameters, we assume that each kernel function has at least one hyperparameter. 
For $n_s>1$ sources, this model has $\mathcal{O}(n_s^3)$ scalar hyperparameters from the matrices $\mathbf{W}_\nu$, $\mathcal{O}(n_s)$ scalar noise hyperparameters $\sigma_i$, and, in addition, the hyperparameters of the kernels $k_\nu$. The computational complexity of training this model with a given set of hyperparameters scales as $\mathcal{O}(N^3)$, where $N$ is the total number of data points from all tasks. This makes it expensive and challenging to optimize the hyperparameters of the model.
In the following, we introduce several simplifications from the literature that have fewer hyperparameters and/or better scaling properties.

\paragraph{\DEFmultitasksinglekGP (\multitasksinglekGP)}
A common heuristic is to assume that the source and target functions share a common structure and consider $k_s=k_t$. 
The joint kernel contains only one coregionalization matrix, which reduces the number of scalar hyperparamters to $\mathcal{O}(n_s^2)$, while the computational complexity is the same as \cref{eq:general_kernel}. 

\paragraph{\DEFweightedSourceGP (\weightedSourceGP)}
Another common simplification is to set correlations between different source data sets to zero
\begin{align}\label{linear-combinations}
    [\mathbf{W}_{s}]_{i,j} &= \delta_{i,s}\delta_{j,s} + w_{st}, 
    &&[\mathbf{W}_{t}]_{i,j} = \delta_{i,t}\delta_{j,t}.
\end{align}
The hyperparameters $w_{st}$ quantify how much the source is correlated with the target. 
Each coregionalization matrix has at most one parameter, which reduces the total parameter number to $\mathcal{O}(n_s)$.

\paragraph{\DEFhierarchicalGP (\hierarchicalGP)}
The asymmetric setting in transfer learning, i.e., modelling the target but not the source tasks, motivates a hierarchical approach to model the differences of target to source data with an additive kernel defined by 
\begin{align}
\label{eq:hierarchical-kernel}
    [\mathbf{W}_{s}]_{i,j} &= 1, 
    &&[\mathbf{W}_{t}]_{i,j} = \delta_{i,t}\delta_{j,t}.
\end{align}
As in \weightedSourceGP, the number of scalar parameters is of order $\mathcal{O}(n_s).$
If optimized jointly, the computational complexity is the same as of \multitaskGP. 

\paragraph{\DEFstackGP (\stackGP)}
\citet{golovin2017stackgp} propose a simpler way to transfer information from a pretrained source model to the target and only use the posterior mean of the source model as prior mean for the target. 
All correlations of the source model are neglected, which leads to
\begin{align}
    [\mathbf{W}_{s}]_{i,j} &= \delta_{i,s}\delta_{j,s}
    &&[\mathbf{W}_{t}]_{i,j} = \delta_{i,t}\delta_{j,t}.
\end{align}
The kernel is block diagonal in the different tasks. The training complexity of \stackGP is therefore the complexity of training the target model plus the additional cost of evaluating the source posterior mean at the target points. Note that \citet{golovin2017stackgp} additionally combine source and target uncertainty heuristically, which we do not consider for ease of exposition.

\begin{table*}
  \caption{Computational complexity for one source task. Training the source model has a complexity of $\mathcal{O}(N_s^3)$. Querying the target has a complexity of $\mathcal{O}(N_t^2 + N_s)$ for MHGP and $\mathcal{O}((N_t + N_s)^2)$ for the other techniques. In \cref{ap:complexity_multiple_sources} we discuss the generalization to multiple sources.}
  \label{tab:computational_complexity}
  \begin{center}
  \begin{tabular}{lllll}
    \toprule
    Models & Abbreviation & HPs & Joint HPO & Training of target \\
    \midrule
    \DEFmultitaskGP
    & \multitaskGP
    & $\mathcal{O}(n_s^3)$
    & yes
    & $\mathcal{O}((N_t + N_s)^3)$  \\
    
    \DEFmultitasksinglekGP
    & \multitasksinglekGP
    & $\mathcal{O}(n_s^2)$
    & yes
    & $\mathcal{O}((N_t + N_s)^3)$ \\
    
    \DEFweightedSourceGP 
    & \weightedSourceGP
    & $\mathcal{O}(n_s)$
    & yes
    & $\mathcal{O}((N_t + N_s)^3)$ \\
    
    \DEFhierarchicalGP 
    & \hierarchicalGP
    & $\mathcal{O}(n_s)$
    & yes
    & $\mathcal{O}((N_t + N_s)^3)$ \\
    
    \DEFseqhierarchicalGP
    & \seqhierarchicalGP
    & $\mathcal{O}(n_s)$
    & no
    & $\mathcal{O}(N_t^3 + N_t^2 N_s + N_t N_s^2)$ \\
    
    \DEFboostedGP
    & \boostedGP
    & $\mathcal{O}(n_s)$
    & no
    & $\mathcal{O}(N_t^3 + N_t^2 N_s + N_t N_s^2)$ \\
    
    \DEFstackGP
    & \stackGP
    & $\mathcal{O}(n_s)$
    & no
    & $\mathcal{O}(N_t^3 + N_t N_s)$ \\
    
    \bottomrule
  \end{tabular}
  \end{center}
\end{table*}

\section{INTERMEDIATE-COMPLEXITY TRANSFER MODELS}
\label{sec:intermediate-complexity-models}
The Bayesian methods presented above require an expensive optimization of the joint likelihood. In this section, we introduce our novel models, \DEFseqhierarchicalGP (\seqhierarchicalGP) and \DEFboostedGP (\boostedGP), which leverage the asymmetric setting of transfer learning to lower the complexity.

\paragraph{\DEFseqhierarchicalGP (\seqhierarchicalGP)}
The starting point is the \hierarchicalGP model from \cref{eq:hierarchical-kernel} in which both the source, $p(\theta_s|\mathcal{D})$, and target, $p(\theta_t|\mathcal{D})$, model parameters are influenced by \emph{all} task data, $\mathcal{D}=\mathcal{D}_s\cup\mathcal{D}_t$. Here, the quantities $\theta_s$ and $\theta_t$ parametrize the prior distribution of the GP. The asymmetric nature of transfer learning, in which the target data are accumulated during optimization while the source data remain invariant, motivates the use of a modular Bayesian approach in which the target data do not influence the source model \citep{bayarri2009modularization}. Training such a model involves (i) finding a suitable $p(\theta_s | \mathcal{D}_s)$ by optimizing the \emph{partial} likelihood, $\mathcal{L}_s$, and (ii) finding $p(\theta_t | \mathcal{D})$ by optimizing the total likelihood while keeping $\mathcal{L}_s$ fixed to the value from the first step. This sequential training procedure is the telltale of \seqhierarchicalGP. Under more restrictive assumptions and scope, a similar approach was proposed in \citet{kennedy2000predicting, le2014recursive}. The modular nature of \seqhierarchicalGP leads to a large decrease in training complexity compared to \hierarchicalGP as shown later. The price to pay for the complexity reduction is the inability of the source model to react to the target data, which may become an issue in case $p(\theta_s|D_s)$ is misspecified.

We focus, for concreteness, on hierarchical models in this paper but note that such modularizations, which decrease the training complexity, can, in principle, be conducted for other Bayesian designs from \cref{sec:models}, too. 
Exploring such approaches is left for future work.



\paragraph{\DEFboostedGP (\boostedGP)}
In \seqhierarchicalGP, the source model parameters, $\theta_s$, do not depend on the target data but the source posterior distribution does. We propose \boostedGP, which makes also the latter independent of the target data. This ad-hoc assumption takes \boostedGP outside the Bayesian realm but provides an elegant connection to boosting, a well-known approach in machine learning that combines an ensemble of weak learners into a strong one \citep{schapire2003boosting}. We study this connection in \cref{sec:equivalences}.
In contrast to \hierarchicalGP, this leads to an asymmetric treatment of observed target data and query points. 
In \cref{sec:boostedgp} we show that this approach results in the kernel
\begin{equation}
\label{eq:boosted_kernel}
    k((\x, i), (\x', j)) = \sum_{r} [\mathbf{W}_\nu]_{i, j} k_\nu(\x, \x') + 
    \delta_{xx'}\delta_{ij}\sigma_i^2
\end{equation}
with $i, j, r \in \{s, t, *\}$, $k_* = k_t + \Sigma_*^{\spl}$, $\sigma_* = 0$, and
\begin{align*}
 [\mathbf{W}_*]_{ij} &= \delta_{i*} \delta_{j*},
 &&[\mathbf{W}_{s}]_{ij} = \delta_{is}\delta_{js},
 &&[\mathbf{W}_{t}]_{ij} = \delta_{it}\delta_{jt}.
\end{align*}
Note that \boostedGP introduces an additive term 
$
\Sigma_*^{\spl}
    = 
    \Sigma^s_{*, *}
    + \woodburry\Sigma^s_{t, t}
    \woodburry^T
    -\woodburry\Sigma^s_{t, *}
    -\Sigma^s_{*, t} \woodburry^T
$
to the covariance of the query points, which resembles a robustified version of the target model. Here $\woodburry = k_t(\x_*, \X_t)\left(k_t(\X_t, \X_t) +\sigma_t^2\mathbf{I}\right)^{-1}$, and $\Sigma_{t,t}^s$, $\Sigma_{t,*}^s$, $\Sigma_{*,t}^s$, $\Sigma_{*,*}^s$ are the blocks of the posterior covariance matrix of the source evaluated at target and query points.
The number of hyperparameters is equal to \stackGP, while the computational complexity is the same as of \seqhierarchicalGP.

\section{THEORETICAL FOUNDATIONS 
}
\label{sec:equivalences}

In \cref{sec:models,sec:intermediate-complexity-models} we introduced several transfer-learning models that rely on the hierarchical kernel architecture. In the following, we present insights on the connection between these methods and their design choices.

\DEFstackGP is the simplest of the models. It trains a GP on the source data and propagates the posterior mean to be the prior mean of the target model.
This approach has the lowest computational complexity, 
see \cref{tab:computational_complexity} (\stackGP).
Neglecting the transfer of uncertainty from the source to target comes at a cost, as shown in the experimental section (\cref{sec:experiments}), since well-calibrated uncertainty estimates are at the core of BO's sample efficiency and neglecting the uncertainty may be detrimental to the optimization.

So far we have adopted a Bayesian view when presenting the transfer models in \cref{sec:models,sec:intermediate-complexity-models}. This view is, however, not particularly useful for the non-Bayesian nature of \boostedGP. We therefore switch to an alternative and unifying view in which uncertainty propagates via the stochastic realizations of the models. In particular, we study the creation of an ensemble of target models based on functional samples from the source posterior. Averaging over the ensemble would then lead to the desired target model.
In this section, we show that both our proposed transfer-learning models, \seqhierarchicalGP and \boostedGP, are closed-form solutions to such averaging procedures.

Performing model averaging over the ensemble of target \emph{priors} results in the target model of \DEFhierarchicalGP, as we show in \cref{sec:bayesian_averaging}. In light of the ensemble averaging, this kernel naturally lends itself to a sequential optimization of hyperparameters resulting in our proposed \DEFseqhierarchicalGP. The key advantages of this approach are the accurate uncertainty estimates with reduced computational complexity, see \cref{tab:computational_complexity} (\hierarchicalGP vs \seqhierarchicalGP), and the sequential optimization of weakly correlated subsets of hyperparameters, which stabilizes training and may improve performance, see \cref{fig:experiments_multi_source}.

By contrast, averaging over the ensemble of target \emph{posterior distributions} leads to \DEFboostedGP, as we show in \cref{sec:models}.
Here, each target posterior in the ensemble is a GP with a sample from the source posterior as prior mean function. 
This model inherits the uncertainty in a non-Bayesian fashion and is fundamentally different from \DEFseqhierarchicalGP.

\begin{figure}[t]
\centering
  \includegraphics[width=\columnwidth]{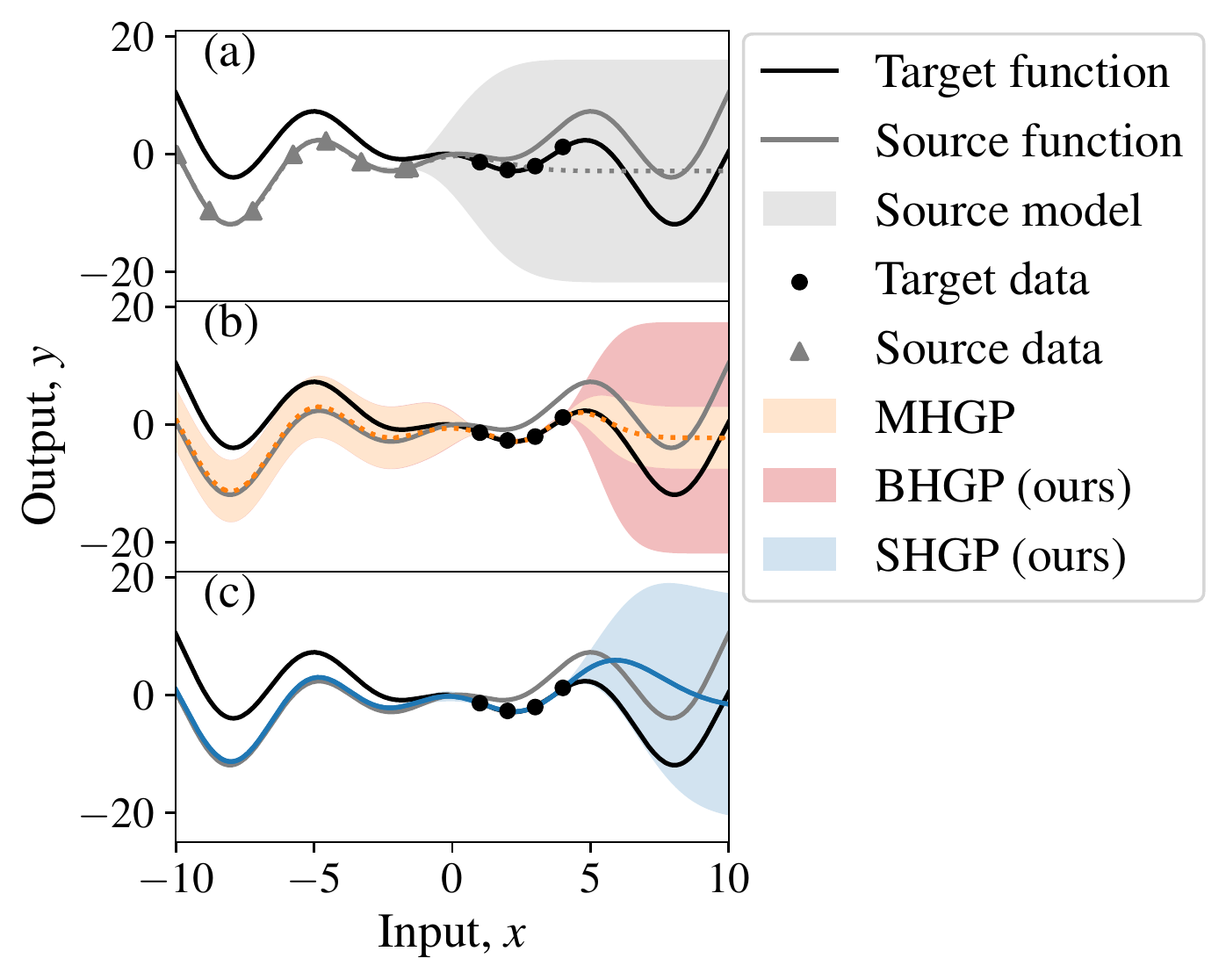}
  \caption{Visualization of key hierarchical transfer-learning models presented in the paper.  
  The posteriors of the source model (a), of \stackGP\ and \boostedGP\ (b), and of \seqhierarchicalGP\ (c) are shown in terms of the mean function and 95\% confidence intervals.}
  \label{fig:comparison}
\end{figure}

Before diving into the theoretical analysis, we illustrate the fundamental differences between the three hierarchical models in \cref{fig:comparison}. In the plots, we train the techniques on data generated from an Alpine function family, $f(x; c)=x\sin(x+\pi) + cx$, with one input, $x\in(-10, 10)$, and one parameter defining the family, $c\in\R$. The source data are generated uniformly within $x\in(-10,0)$ with $c=1/2$, which is why the source posterior distribution is uncertain on the right, $x\in(0,10)$, see \cref{fig:comparison}(a). The target data are generated with $c=-1/2$ for $X_t=(1, 2, 3, 4)$.
The posterior of \DEFstackGP\ underestimates the uncertainty of the target function on the right, since its uncertainty originates solely from the target data, see \cref{fig:comparison}(b). \DEFboostedGP alleviates this shortcoming on the right, while on the left it has the same uncertainty since the source model is confident. The posterior of \DEFseqhierarchicalGP is Bayesian and correctly captures the variation of the target function on the right, see \cref{fig:comparison}(c). On the left, \DEFseqhierarchicalGP follows the source model, since the target points are well explained by the source model alone, see \cref{fig:comparison}(a), and no extra uncertainty is required.

\subsection{Bayesian model averaging}\label{sec:bayesian_averaging}
In this section, we show that Bayesian model averaging over prior mean functions of the target, which are distributed according to the posterior of the source, leads to the \hierarchicalGP model defined by \cref{eq:hierarchical-kernel}.

\begin{restatable}{proposition}{claimuikernel}\label{claim:ui_kernel}
Let $f_s \sim \GP{0}{k_s}$, $f_t|f_s\sim \GP{f_s}{k_t}$, and 
\begin{equation}
    \label{eq:bayesianaverage}
    p(\uidist{t}{} \mid \mathcal{D}_t)=\int p(f_t \mid f_s, \mathcal{D}_t) p(f_s \mid \mathcal{D}_t) df_s.
\end{equation}
Then, the joint model of $\funcdist{}{s}{}$ and $\uidist{t}{}$ is a GP with zero mean prior and the kernel \cref{eq:hierarchical-kernel}.
\end{restatable}

We provide a closed form solution for the marginalization of \cref{eq:bayesianaverage} over the source posterior in \cref{ap:equivalences}. Since both integrands are Gaussian, the integral yields a Gaussian distribution for $p(\uidist{t}{} \mid \mathcal{D}_t)$.
By conditioning the joint model from
\cref{claim:ui_kernel} on the source data, we can derive the following prior for the target model.
\begin{restatable}{corollary}{claimuiprior}
\label{claim:uiprior}
Let 
$
k_\Sigma 
    = k_t + \cov{f_s \mid \mathcal D_s}.
$
Then under the assumptions of \cref{claim:ui_kernel} it holds that $\uidist{t}\sim\ \GP{\expectation{f_s \mid \mathcal D_s}}{k_\Sigma}.$
\end{restatable}
We provide the detailed derivations in \cref{ap:equivalences}. 
The training complexity scales as $\mathcal{O}(N_t^3 + N_t^2 N_s + N_t N_s^2)$, see \cref{ap:subsec:complexity-shgp} for the proof. We therefore save, during training, the steep complexity contribution of $\mathcal{O}(N_s^3)$ of the conventional Bayesian methods.

\subsection{Posterior prediction averaging}\label{sec:boostedgp}
In an alternative approach we take inspiration from the well-known principle of boosting \citep{schapire2003boosting} and average over the posterior distributions of the target models.
In contrast to Bayesian model averaging in \DEFseqhierarchicalGP, this approach neglects the implicit dependency of the source model on the target data, $p(f_s\mid\mathcal{D}_t)\rightarrow p(f_s)$ in~\cref{eq:bayesianaverage}. This distribution is analytically tractable and equals the posterior distribution of a GP with kernel~\cref{eq:boosted_kernel}, i.e., the \DEFboostedGP. 

\begin{restatable}{proposition}{claimboosting}
\label{claim:boosting}
Let $f_s, f_t$ be as in \cref{claim:ui_kernel} and
\begin{equation*}
    p(\spldist{t}{} \mid \mathcal{D}_t)=\int p(f_t \mid f_s, \mathcal{D}_t) p(f_s) df_s.
\end{equation*}
Then, the joint model of $\funcdist{}{s}{}$ and $\spldist{t}{}$ is a GP with zero mean and the covariance function defined in \cref{eq:boosted_kernel}.
Further, $\spldist{t}{}\mid\x_*, \mathcal D_t$ is multi-variate normal with mean
$\expectation{f_s\mid\x_*, \mathcal D_s} - \woodburry(\mathbf{y}_t-\expectation{f_s\mid\X_t, \mathcal D_s})$ and covariance matrix $k_t(\x_*, \x_*) -\woodburry k_t(\X_t,\x_*)+\Sigma_*^{\spl}.$
\end{restatable}

The proof follows the same ideas as for \cref{claim:uiprior} and can be found in \cref{ap:equivalences}. Since the prior for the target of \boostedGP coincides with the prior of \stackGP, the hyperparameter optimization is the same. 
The training complexity is the same as for \seqhierarchicalGP, $\mathcal{O}(N_t^3 + N_t^2 N_s + N_t N_s^2)$, see \cref{ap:subsec:complexity-bhgp} for proof.

\section{EXPERIMENTS}\label{sec:experiments}
We provide an experimental evaluation of the transfer-learning algorithms discussed in \cref{tab:computational_complexity}. In addition, we evaluate three baselines: GP-based BO (GPBO) without any source data, RGPE \citep{feurer2018rgpe} in which the target function is modelled as a weighted sum of the predictions of all task GPs, and GC3P \citep{salinas2020quantile} in which a Gaussian Copula regression with a parametric prior is employed to scale to large data. We implement the models using GPy \citep{gpy2014} and run BO with Emukit \citep{emukit2019}, licensed under BSD 3 and Apache 2.0, respectively. For GC3P we use the publicly available code~\citep{salinas2020quantile}. GP hyperparameters are optimized by maximizing the likelihood for the observed data. We focus on maximum likelihood for simplicity but, in high dimensions, it may be preferable to employ a Bayesian treatment of hyperparameters. Code to reproduce our results can be found on Github\footnote{\url{https://github.com/boschresearch/transfergpbo}}.

\subsection{Synthetic Function Families}
The synthetic function families that are derived from conventional benchmark functions by placing probability distributions on their parameters. We consider a broad spectrum of families, where the optimum of tasks varies locally for some and in the entire domain for others: the one-dimensional Alpine \citep{momin2013literature} function, and the multi-dimensional Branin, Hartmann3 and Hartmann6. Details about the function families are provided in \cref{ap:synthetic_functions}. Performance is reported via mean simple regret over 50 runs together with standard error of the mean. 

\begin{figure*}
  \includegraphics[width=\textwidth]{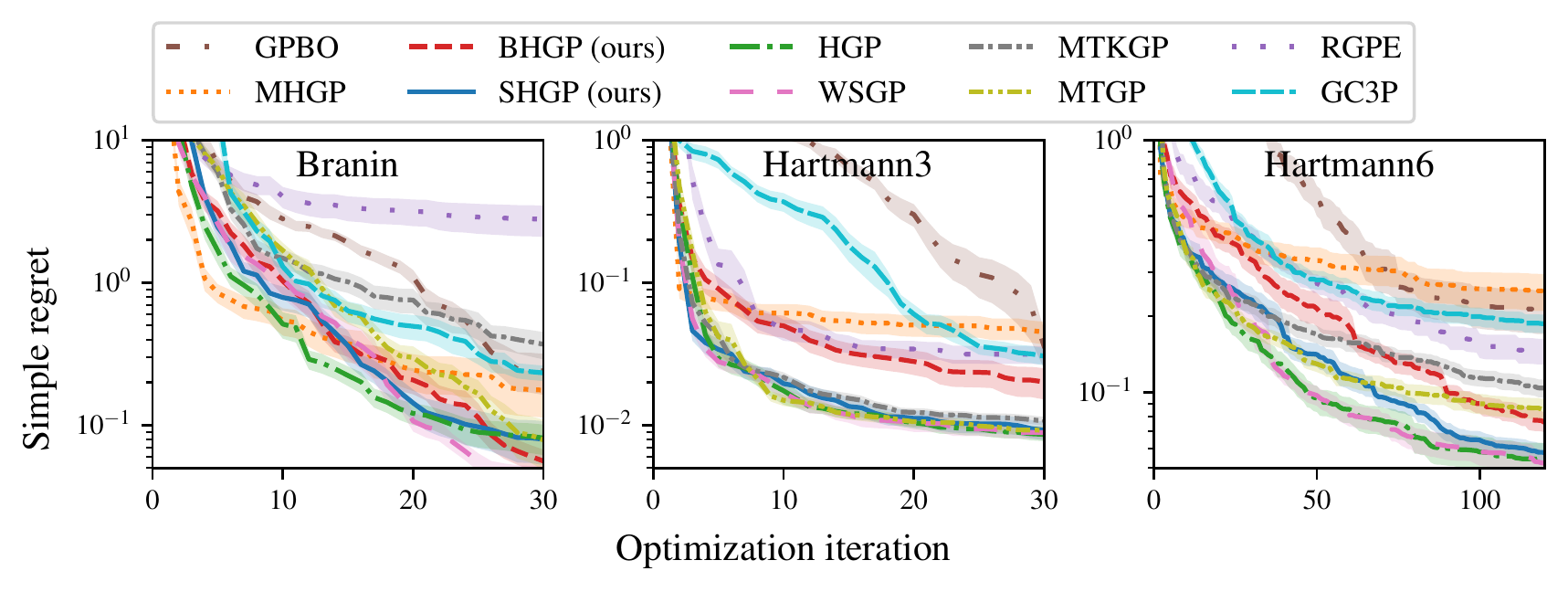}
  \caption{Performance on the single-source benchmarks: the two-dimensional Branin (left), three-dimensional Hartmann3 (middle), and six-dimensional Hartmann6 (right). The source data are sampled randomly from the source function and contain $20N_\mathrm{dim}$ points with $N_\mathrm{dim}$ being the input dimension of the benchmark. We add i.i.d. observational noise of standard deviation $\sigma_s = \sigma_t = 0.1$ for Hartmann3, Hartmann6, and $\sigma_s = \sigma_t = 1.0$ for Branin during data generation.
  }
  \label{fig:experiments_single_source}
\end{figure*}

\paragraph{Single-source benchmark functions}\label{sec:single_source_experiments} We start the experimental analysis in the simplest setting with a single source task. This fundamental regime provides valuable insight into the trade-off between performance and scalability of the transfer learning algorithms. Applications with scarce historical data where the transfer efficiency is important may particularly benefit from this analysis.

The algorithms are benchmarked on three function families, see \cref{fig:experiments_single_source}. Results on two further function families are available in \cref{ap:results_1d_benchmarks}. A notable difference in the convergence on the different benchmarks can be observed. This is likely to be caused by the different particularities of each function family, which we discuss in \cref{ap:synthetic_functions}. Despite these differences, there are a few important generalizing features that characterize the algorithms:
(i) The Bayesian techniques \seqhierarchicalGP, \hierarchicalGP, and \weightedSourceGP outperform the non-Bayesian algorithms. This is not surprising, since Bayesian models compute highly informed probabilistic predictions. The best and most consistent performers are the \hierarchicalGP and \weightedSourceGP algorithms. We attribute this to the flexibility of their design. During training, all kernel's hyperparameters are jointly optimized allowing for superior model quality.
(ii) Our modular Bayesian technique, \seqhierarchicalGP, is competitive and an attractive alternative with lower computational complexity. The other non-Bayesian methods trail behind, which is likely due to their inability to reliably propagate model uncertainty. Among these, our technique, \boostedGP, performs consistently better than other non-Bayesian techniques and provides a reasonable compromise between computational complexity and efficiency.
(iii) Surprisingly, the most advanced and expensive techniques, \multitasksinglekGP and \multitaskGP, perform worse than the other Bayesian techniques. This is likely caused by the challenging training procedure, where a large number of hyperparameters are optimized.

\begin{figure*}
  \centering
  \includegraphics[width=\textwidth]{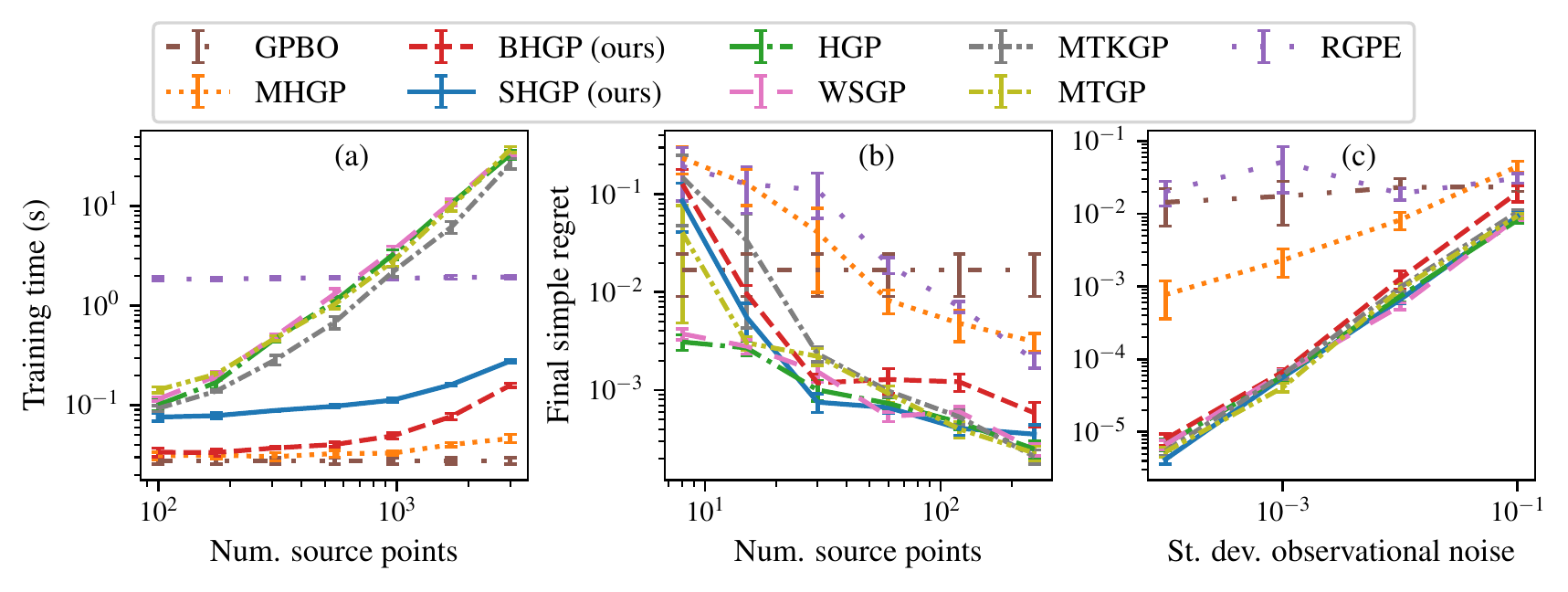}
  \caption{Runtime analysis (a) and the impact of the size of the source data set (b) and observational noise (c) on the algorithm performance. In (a) we plot the training time in seconds versus number of points in the source data set for the Hartmann6 function family. We consider the timing of one single step of gradient ascent during the optimization of the likelihood function. The target data set is set to 100 randomly sampled points. Statistics are acquired via 7 independent runs. The final simple regret versus number of points for constant observational noise, $\sigma_s = \sigma_t = 0.01$ (b), and standard deviation of observational noise for 60 source points (c), is plotted for the Hartmann3 function family. The performance of GPBO is independent of the number of historical points. }
  \label{fig:experiments_num_source_and_noise}
\end{figure*}

In addition to the asymptotic complexities in \cref{tab:computational_complexity}, we demonstrate the lower computational complexity of our methods in a direct runtime analysis of model training in \cref{fig:experiments_num_source_and_noise}(a). \seqhierarchicalGP and \boostedGP are orders of magnitude faster than the full Bayesian methods for moderately large source data sets. The runtimes reported in \cref{fig:experiments_num_source_and_noise}(a) are representative of the entire optimization runtime since model training has steeper complexity than acquisition-function optimization, see \cref{ap:runtime_analysis} for empirical evidence.

\paragraph{Observational noise and propagation of uncertainty}\label{subsec:exp_impact_noise} Beyond these generic trends, the performance of the techniques depends on other factors like the amount of source data and the magnitude of the observational noise. Such a study is presented in \cref{fig:experiments_num_source_and_noise}(b, c); more results are available in \cref{ap:ablation_studies}. We distinguish three data regimes:
(i) In the limit of scarce source data, which is insufficient to describe the global shape of the source function, the kernel methods with joint HPO, \hierarchicalGP and \weightedSourceGP, outperform the other methods significantly. The reason is that they model all data jointly, in contrast to the methods trained sequentially that end up with source models of poor quality.
(ii) The case of moderate amount of source data, which is sufficient to describe the source function probabilistically but not deterministically, is discussed in the previous section.
(iii) In the limit of lots of source data, which is sufficient to deterministically describe the function, all transfer-learning methods converge to a similar performance. Here, propagation of uncertainty is not required and the non-Bayesian approaches are appealing due to their favorable scaling.

The observational noise affects the boundary between the aforementioned data regimes. Enhanced noise leads to an increased amount of data required by a model of fixed quality.

\begin{figure*}
  \centering
  \includegraphics[width=\textwidth]{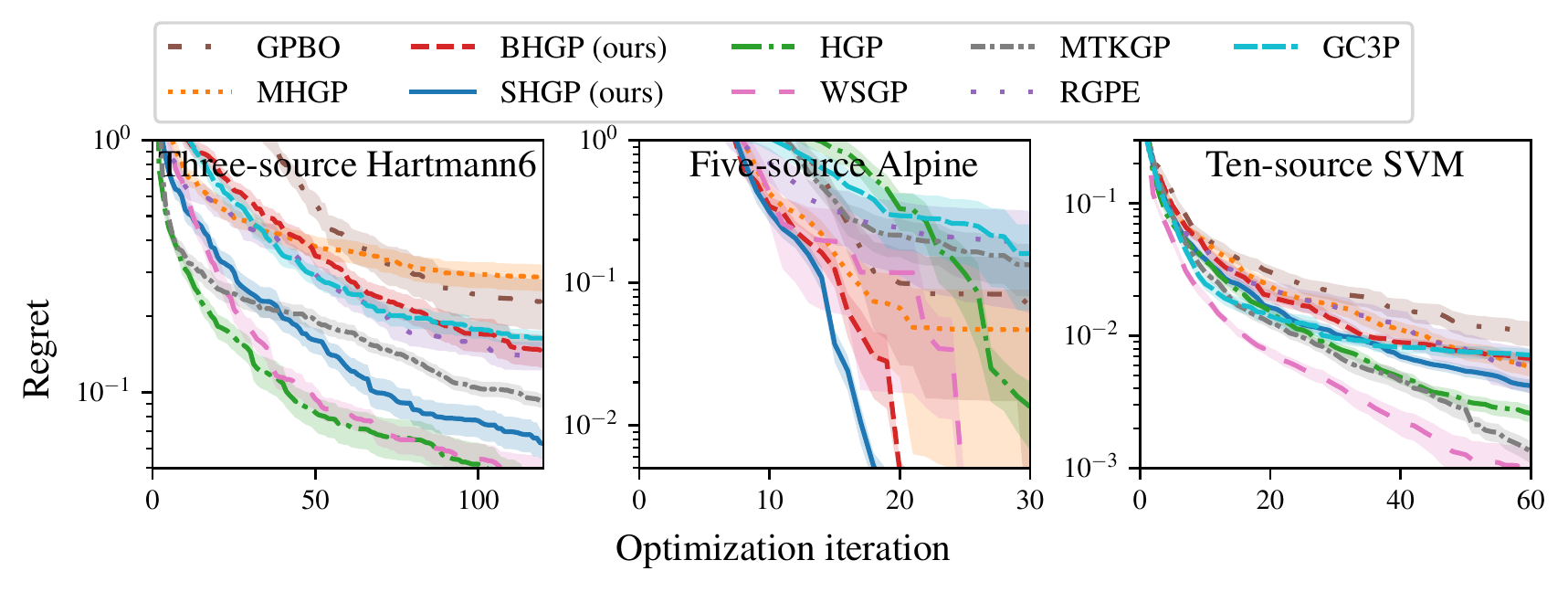}
  \caption{Performance on the multi-source benchmarks: the three-source Hartmann6 (left), five-source Alpine (middle), and ten-source OpenML SVM benchmark (right). The source data are sampled randomly and contain 60, 20, and 60 points per task for Hartmann6, Alpine, and SVM, respectively. I.i.d. observational noise of standard deviation $\sigma_s = \sigma_t = 0.1$  is added to the source and target data of the synthetic benchmarks.}
  \label{fig:experiments_multi_source}
\end{figure*}

\paragraph{Multi-source benchmark functions}\label{sec:multi_source_experiments} For the more general case of multiple source data sets, the algorithms are benchmarked on two function families: (i) the six-dimensional Hartmann6 family with three source data sets in which the functions are sampled randomly as in \cref{sec:single_source_experiments}, and (ii) the one-dimensional Alpine family with five source data sets, where the source and target functions are fixed as in \citep{feurer2018rgpe}. \multitaskGP is not benchmarked here because of its complexity. The performance of the algorithms on the three-source Hartmann6 is consistent with the single-source benchmarks, compare \cref{fig:experiments_single_source,fig:experiments_multi_source}. \hierarchicalGP and \weightedSourceGP still perform best, while our less complex \seqhierarchicalGP provides a competitive compromise. The picture changes in the Alpine benchmark, where the more structured approaches to HPO show superior performance. Our techniques, \seqhierarchicalGP and \boostedGP, outperform others despite the lower computational complexity. This is likely due to the stable, hierarchical training procedure, where at most three hyperparemeters are optimized at a time. By contrast, \hierarchicalGP optimizes 18 hyperparameters jointly, \weightedSourceGP 23, and \multitasksinglekGP 33. Algorithms employing hierarchical hyperparameter optimization are therefore ever more appealing for increasing number of source tasks.

\subsection{Meta-Learning Surrogate Benchmarks} In the following we study the performance of the transfer-learning techniques on hyperparameter optimization problems in machine learning. We follow \cite{perrone2018scalable} and use evaluations of support vector machine (SVM) models on 28 data sets. The data were published by \citet{kuhn2018automatic} on the OpenML platform \citep{vanRijn2013Collaborative} under the creative commons license \citep{kuhn2018openml}. More details are available in \cref{app:meta-learning-benchmarks}. As in \cite{salinas2020quantile}, we carry out a discrete optimization at the observed evaluations to avoid potential problems related to the bias of a fitted surrogate model.

Since the data sets contain over fifty thousand evaluations, which is too much for the closed-form Bayesian techniques to handle, we perform downsampling and choose 11 tasks at random. The data of the target task are used as the ground truth for the discrete optimization, while the data of the source tasks are downsampled to 60 points per task. We present results averaged over 500 independent runs. For better comparability, we employ a rescaled version of the regret function called \emph{average distance to the global minimum} \citep{Wistuba}, where the regret values on each data set are rescaled between zero and one \emph{before} the averaging procedure. The results are shown in the right panel of \cref{fig:experiments_multi_source}.

Despite the markedly different nature of the data compared to the synthetic benchmarks, the performance of the different transfer-learning techniques exhibits somewhat similar trends. The performance is dominated by \weightedSourceGP, which achieves achieve a regret of 0.01 more than three times faster than classical GPBO. Interestingly, \hierarchicalGP performs worse. We hypothesize this to be caused by the incompatibility between the hierarchical nature of \hierarchicalGP and the structure of the SVM data set. Our techniques, \seqhierarchicalGP and \boostedGP, provide, as before, a tradeoff between complexity and performance, and achieve the same regret level about twice as fast as GPBO.

\section{DISCUSSION AND CONCLUSION}\label{sec:conclusion}
In this paper, 
we have presented a unified view on transfer learning methods based on Gaussian processes with tractable model posteriors. One end of this spectrum is populated by models that maintain a full covariance matrix of all source and target data and perform proper Bayesian inference on all data jointly. While these are powerful models with well-calibrated uncertainty estimates, they quickly become computationally infeasible due to their cubic complexity in the cumulated number of source and target data points. The other side of the spectrum is occupied by heuristics that aggregate the predictions of separate, independently trained, task models. These methods exhibit a much more favourable computational complexity, but unfortunately fail to properly propagate uncertainty between the individual task models. To amend this unsatisfactory binary choice between computational feasibility and uncertainty propagation, we proposed two novel transfer learning models, \DEFseqhierarchicalGP and \DEFboostedGP, that both provide a compromise between these extremes: they add some computational complexity to account for the uncertainty of the involved models with some slight approximations as compared to the full covariance approaches.
Our analysis is supported by comprehensive experiments that pinpoint strengths, weaknesses, and trade-offs of these transfer learning methods.
The benchmarks demonstrate the appeal of \DEFseqhierarchicalGP and \DEFboostedGP as robust and competitive techniques.

\subsubsection*{Limitations}
The low-data regime is challenging for any method. In particular, the empirical success of Gaussian processes hinges on the smoothness assumptions encoded in the kernel. At the same time, it is not obvious how non-smooth functions can be optimized efficiently. Another limitation is that they do not scale to large scale data without approximations, which is not a problem in our setting since we specifically focus on the low-data regime. We are not aware of any negative societal impacts of our work, since we focus on improving the data efficiency of optimization methods.

\subsubsection*{Acknowledgements}
We would like to thank Stefan Falkner for insightful discussions on the results.

\clearpage
\bibliographystyle{plainnat}
\bibliography{references}


\clearpage
\appendix

\thispagestyle{empty}

\onecolumn \makesupplementtitle



\part{Overview} 

In the appendix we provide the detailed proofs for all claims in the paper, complexity analysis for all discussed methods for multiple sources, ablation studies, and the detailed hyperparameter configurations used in the paper. An overview is shown below.

\noptcrule
\parttoc 
\clearpage

\section{EQUIVALENCES}
\label{ap:equivalences}
In the following sections, we provide the proofs of the claims
in the main paper. For ease of reference, we restate those claims here.
We begin with deriving closed form solutions for the prior and posterior of Bayesian model averaging.

\claimuikernel*

\claimuiprior*

Subsequently, we show that posterior prediction averaging leads to a analytically tractable distribution which equals the Boosted Hierarchical GP \cref{eq:boosted_kernel}.

\claimboosting*

The analysis of both ensemble averaging techniques introduced in \cref{sec:equivalences} is based on marginalization of the models over the posterior of the source at target data points $\X_t$ and query points $\x_*.$
Samples from the source posterior can be obtained from the posterior mean $\expectation{f_s \mid \mathcal D_s}$ and covariance of the source $\cov{f_s \mid \mathcal D_s}$ posterior mean and covariance of the source model as follows.

Let 
$L$ be the Cholesky decomposition of posterior covariance of the source $LL^T=\cov{f_s \mid \tilde\X_t, \mathcal D_s}$ evaluated at target data points and query points $\tilde\X_t =\X_t\cup\x_*.$
Let $\varepsilon$ be a standard normal vector of 
size $N_t+1$ then 
\begin{align}
    f_{s,\varepsilon}(\tilde\X_t)&=\expectation{f_s\mid \tilde\X_t, \mathcal D_s}+L_s\varepsilon
\end{align} 
is a sample from the source posterior at target data and query points.
The following commonly known result 
will be useful for deriving the desired Gaussian distributions. For the sake of completeness, we provide a proof of 
\cref{lem:sample_mean} in 
\cref{ap:proof_sample_mean}.
\begin{lemma}\label{lem:sample_mean}
Let $\varepsilon$ be a standard normal random vector of size $N\times 1$. Let $\mu\in\mathbb R^N$, $L\in\mathbb R^{M\times N}, \Sigma\in\mathbb R^{M\times M}$. If  $Y\mid\varepsilon\sim\normaldist{\mu + L\varepsilon}{\Sigma},$ then $Y\sim\normaldist{\mu}{\Sigma+LL^T}$
\end{lemma}

We begin with the proof of \cref{claim:uiprior}, which is a simple application of \cref{lem:sample_mean} to the ensemble of priors used in Bayesian model averaging.
We derive \cref{claim:ui_kernel} from \cref{claim:uiprior} by showing that conditioning the joint model from \cref{eq:hierarchical-kernel} leads to the same posterior.
The proof of \cref{claim:boosting} requires a more involved application of  \cref{lem:sample_mean} to the ensemble of posteriors, that we propose in posterior prediction averaging.
A detailed derivation of \cref{lem:sample_mean} can be found in \cref{ap:proof_sample_mean}.

\subsection{Proof of \cref{claim:uiprior}}
In Bayesian model averaging, we build an ensemble of priors for the target model from multiple functional samples of the source posterior. For a fixed functional sample $f_{s,\varepsilon}(\tilde\X_t)$ the prior of the target is a multi-variate normal with mean $\expectation{f_s \mid \tilde\X_t, \mathcal D_s}+L\varepsilon$ and covariance matrix $k_t(\X_t, \X_t).$ Consequently, \cref{claim:uiprior} follows from \cref{lem:sample_mean} and $L L^{T} =\cov{f_s \mid \tilde\X_t, \mathcal D_s}$.

\subsection{Proof of \cref{claim:ui_kernel}}
To show \cref{claim:ui_kernel}, we start with the model defined in \cref{eq:hierarchical-kernel} and condition it on the source data.
The joint kernel of the Hierarchical GP kernel is given by 
\begin{align}
    k((\x,i),(\x',j)) &= k_s(\x,\x') + \delta_{i,t}\delta_{j,t} k_t(\x,\x')
\end{align}
and after conditioning this model on source data $\mathcal D_s$, we obtain the prior of the target model by standard GP inference. 
Then, 
\begin{align}
k((\tilde\X_t,t),(\X_s,s))&=k_s(\tilde\X_t, \X_s),&
k((\X_s,s),(\tilde\X_t,t))=k_s(\X_s, \tilde\X_t),\\
k((\tilde\X_t,t),(\tilde\X_t,t))&=k_s(\tilde\X_t, \tilde\X_t)+k_t(\tilde\X_t, \tilde\X_t),&
k((\X_s,s),(\X_s,s))=k_s(\X_s,\X_s).
\end{align}
As a consequence, the posterior of the Hierarchical GP model given the source data has mean and covariance
\begin{align}
    \mu^\ui
    &= k_s(\tilde\X_t, \X_s)[k_s(\X_s,\X_s)]^{-1}y_s,\\
    \Sigma^\ui
    &=k_s(\tilde\X_t,\tilde\X_t)+k_t(\tilde\X_t, \tilde\X_t) + k_s(\tilde\X_t,\X_s)[k_s(\X_s, \X_s)]^{-1}k_s(\X_s,\tilde\X_t).
\end{align}
We can rewrite
\begin{align}
\mu^\ui &=\expectation{f_s \mid \tilde\X_t, \mathcal D_s}
\\
\Sigma^\ui&=k_t(\tilde\X_t, \tilde\X_t) + \cov{f_s \mid \tilde\X_t, \mathcal D_s}.
\end{align}
Together with \cref{claim:uiprior}, we obtain that conditioning the Hierarchical GP model defined in \cref{eq:hierarchical-kernel} on source data leads to the prior target model of Bayesian model averaging.
\cref{claim:ui_kernel} follows because under the same modelling assumption for the source, the prior is unique.

\subsection{Proof of \cref{claim:boosting}}
In this subsection, we compute the distribution obtained from directly averaging over the posteriors on the target data induced by the mean priors sampled from the source posterior (and, thus, ignoring the implicit dependency of the source model on the target data).
Denote by $L_{s,*}$ and $L_{s,t}$ the first $N_*$ and remaining $N_t$ rows of $L_s.$
For a fixed sample $f_{s,\varepsilon}(\X_t, \x_*)$, the posterior of the target GP at $\x_*$ is a multivariate normal with mean and covariance matrix
\begin{align*}
    \mu^\spl &= \expectation{f_s \mid \x_*, \mathcal D_s}+L_{s,*}\varepsilon
    +\woodburry\left[y_t-\expectation{f_s \mid \X_t, \mathcal D_s}-L_{s,t}\varepsilon\right],\\
    \Sigma^\spl &= k_t(\x_*,\x_*) - \woodburry k_t(\X_t, \x_*).
\end{align*}
The posterior mean can be rewritten as 
\begin{align}
    \mu^\spl &=  \expectation{f_s \mid\x_*, \mathcal D_s}
    +\woodburry\left[y_t-\expectation{f_s\mid \X_t, \mathcal D_s}\right]
    +L^\spl\varepsilon
\end{align}
with 
$L^\spl=L_{s,*}+\woodburry L_{s,t}.$
Recall, that $\Sigma_{t,t}^s$, $\Sigma_{t,*}^s$, $\Sigma_{*,t}^s$, $\Sigma_{*,*}^s$ denote the blocks of the posterior covariance matrix of the source evaluated at target and query points.
Then it holds that 
\begin{align}
L_{s,*}L_{s,*}^T &= \Sigma_{*,*}^s,&
L_{s,t}L_{s,t}^T = \Sigma_{t,t}^s,\\
L_{s,*}L_{s,t}^T &= \Sigma_{*,t}^s,&
L_{s,t}L_{s,*}^T = \Sigma_{t,*}^s.
\end{align}
From this, we obtain
\begin{align}
    L^\spl (L^{\spl})^T = 
    \Sigma^s_{*, *}
    + \woodburry\Sigma^s_{t, t}
    \woodburry^T
    -\woodburry\Sigma^s_{t, *}
    -\Sigma^s_{*, t} \woodburry^T=\Sigma_*^{\spl}.
\end{align}
With this observation, the posterior distribution stated in \cref{claim:boosting} can be derived from \cref{lem:sample_mean}.

Finally, it remains to show that the joint model defined in \cref{eq:boosted_kernel} leads to the same posterior distribution. 
Since the source kernel is only evaluated at source points, conditioning on the source first yields
\begin{align}
\label{eq:sampling-method-prior}
\begin{pmatrix}
\spldist{t}{(\x_*)}\\
\mathbf{y}_t
\end{pmatrix}
    &\sim 
    \normaldist{
    \begin{pmatrix}
    \expectation{f_s\mid\x_*, \mathcal D_s}\\
    \expectation{f_s\mid\X_t, \mathcal D_s}
    \end{pmatrix}
        }{
    \begin{pmatrix}
    k_t(\x_*, \x_*) + \Sigma_*^{\spl} 
    & k_t(\x_*, \X_t) \\
    k_t(\X_t, \x_*) & k_t(\X_t, \X_t) + \sigma_t^2\mathbf{I}
    \end{pmatrix}}.
\end{align}
Note that the additive term $\Sigma_*^\spl$ only contributes to the covariance at query points. 
That is, it will only contribute an additive term to the posterior covariance.
Using standard formulas for conditional distributions in multi-variate Gaussians, we obtain that $\spldist{t}{(\x_*)}|\mathbf{y}_t$ is multi-variate normally distributed with mean $\expectation{f_s\mid\x_*, \mathcal D_s} - \woodburry(\mathbf{y}_t-\expectation{f_s\mid\X_t, \mathcal D_s})$
and covariance matrix
$
k_t(\x_*, \x_*) -\woodburry k_t(\X_t,\x_*)+\Sigma_*^{\spl}.
$

\subsection{Proof of \cref{lem:sample_mean}}\label{ap:proof_sample_mean}
Let
$\normalPDF{x}{\mu+L\varepsilon}{\Sigma}$ denote the density of a multi-variate normal distribution with mean $\mu+L\varepsilon$ and covariance matrix $\Sigma$.
Further, denote by $\abs{\Sigma}$ the determinant of $\Sigma.$ 
Then, 
\begin{align}
\normalPDF{x}{\mu+L\varepsilon}{\Sigma}
&= \left(2\pi\right)^{-N/2}
\abs{\Sigma}^{-1/2}
\exp\left[-\frac{1}{2}\left[x-\mu - L\varepsilon\right]^T \Sigma^{-1}\left[x-\mu + L\varepsilon\right]\right]\\
&=
\abs{ L^TL}^{-1/2} 
\normalPDF{\varepsilon}{L^{-1}(x-\mu)}{L^{-1}\Sigma \left(L^{-1}\right)^T}\label{eq:sample_mean}
\end{align}
Similarly as above, denote by 
$\normalPDF{x}{0}{\mathbb 1}$ denote the density of a spherical multi-variate Gaussian.
Then from \cref{eq:sample_mean}, we obtain
\begin{align}
&\normalPDF{x}{\mu+L\varepsilon}{\Sigma}
\normalPDF{\varepsilon}{0}{\mathbf 1}\nonumber\\
&\quad=
\normalPDF{x}{\mu}{\Sigma+LL^T}
\normalPDF{\varepsilon}
{L^{-1}(L^T+\Sigma^{-1})(x-\mu)}{\mathbb 1 + L^{-1}\Sigma \left(L^{-1}\right)^T}.
\end{align}
From this the result follows by integrating out $\varepsilon$
\begin{align}
\int
\normalPDF{x}{\mu+L\varepsilon}{\Sigma}
\normalPDF{\varepsilon}{0}{\mathbf 1}d\varepsilon
&= \normalPDF{x}{\mu}{\Sigma+LL^T}.
\end{align}

\newpage
\section{GENERALIZATION TO MULTIPLE SOURCES}
\label{ap:complexity_multiple_sources}
In this appendix, we discuss the generalization of \cref{sec:models} to multiple sources. We discuss the computational complexity to predict mean and covariance of the target model at a new query point $x_*$. We collect all calculations that can be done without knowledge of the query point as ``training complexity'' and the remainder as ``prediction complexity''. In this section we assume $n_s>1$ sources. For easier generalization we denote the sources by $\nu=1, 2, ..., n_s$ and the target by $\nu=n_s+1$. Where appropriate, the analysis is structured in four parts: Training of the first source, training of the $n>1$th source assuming all ``lower'' sources $n'<n$ have already been trained, training of the target assuming all sources have been trained, and prediction from the target model. Our basis operation for evaluating the computational complexity is scalar multiplication. Sometimes, we will also present our results using $N_t=N_{n_s+1}$ and $N_s=\sum_{\nu=1}^{n_s} N_\nu$.

\subsection{\DEFmultitaskGP (\multitaskGP)} 
The joint kernel for $n_s$ sources and the target in \DEFmultitaskGP between two points $(\x, i), (\x', j)$ from tasks $i,j \in \{ 1, 2, ..., n_s+1 \}$ is a sum of separable kernels given by
\begin{equation}
\label{eq:general_kernel-general}
    k(( \x, i), ( \x', j)) = \sum_{\nu=1}^{n_s+1} [\mathbf{W}_\nu]_{i, j} \, k_\nu( \x, \x') + 
    \delta_{\x \x'}\delta_{ij}\sigma_i^2.
\end{equation}
The number of hyperparameters is $\mathcal{O}(n_s^3)$. Training and prediction complexity are $\mathcal{O}((N_t+N_s)^3)$ and $\mathcal{O}((N_t+N_s)^2)$, as for a simple GP with $N_t+N_s$ data points. 

\subsection{\DEFmultitasksinglekGP (\multitasksinglekGP)}
The only difference to \DEFmultitaskGP is that $k_\nu = k_s$, which reduces the number of hyperparameters to $\mathcal{O}(n_s^2)$.

\subsection{\DEFweightedSourceGP (\weightedSourceGP)}
The joint kernel for $n_s$ sources and the target in \DEFweightedSourceGP is given by
\begin{align}\label{linear-combinations-general}
    [\mathbf{W}_{\nu}]_{i,j} &= \delta_{i,\nu}\delta_{j,\nu} + w_{\nu} \left(\delta_{i,\nu}+\delta_{i,n_s+1}\right)\left(\delta_{j,\nu} + \delta_{j,n_s+1}\right) \text{ for } \nu=\{1, 2, ..., n_s\}, \\  
    [\mathbf{W}_{n_s+1}]_{i,j} &= \delta_{i,n_s+1}\delta_{j,n_s+1}.
\end{align}
Since all the blocks between different sources are zero, inverting the kernel matrix $K$ is computationally simpler than for the more general kernels. We define the source source-block (which is diagonal in $i,j$) $A=[K]_{i\leq n_s, j\leq n_s}$, the target block $D=[K]_{n_s+1, n_s+1}$, and the connecting blocks $B=[K]_{i\leq n_s, j=n_s+1}$ and $C=[K]_{i=n_s+1, j\leq n_s}$. With this definition, we see that using block-inversion \citep{blockinversion}.
\begin{align}
K^{-1}
    &=
    \begin{pmatrix}
        A & B \\
        B^T & D
    \end{pmatrix}^{-1}
    =
    \begin{pmatrix}
        A^{-1} + A^{-1} B (D - B^T A^{-1} B)^{-1} B^T A^{-1} & -A^{-1} B (D - B^T A^{-1} B)^{-1} \\
        -(D - B^T A^{-1} B)^{-1} B^T A^{-1} & (D - B^T A^{-1} B)^{-1}
    \end{pmatrix}
\end{align}
we can use the block-diagonal structure of $A$ for an efficient inversion of the kernel matrix. The total complexity is $\mathcal{O}(N_t^3 + N_t^2 N_s + N_t N_s^2 + \sum_{\nu=1}^{n_s} N_\nu^3).$ Note how due to the blockdiagonal structure of A, the term $(\sum_{\nu=1}^{n_s} N_\nu)^3$ for the inversion of a matrix of the size of A collapses to the much more favourable $\sum_{\nu=1}^{n_s} N_\nu^3$. Prediction at a query point
is of order $\mathcal{O}((N_t+N_s)^2)$.

\subsection{\DEFhierarchicalGP (\hierarchicalGP)}
The joint kernel for $n_s$ sources and the target in \hierarchicalGP is defined by \cref{eq:general_kernel-general} with
\begin{align}
\label{eq:hierarchical-kernel-general}
    [\mathbf{W}_{\nu}]_{i\geq \nu, j\geq \nu} &= 1 \text{ and } 0 \text{ otherwise }.
\end{align}
The number of hyperparameters is hence the number of hyperparameters in the kernel functions, which is of order $\mathcal{O}(n_s)$. When optimizing the hyperparameters of all sources and the target jointly, the training and prediction complexity are $\mathcal{O}((N_t+N_s)^3)$ and $\mathcal{O}((N_t+N_s)^2)$, as for a simple GP with $N_t+N_s$ data points. 

\subsection{\DEFseqhierarchicalGP (\seqhierarchicalGP)}
\label{ap:subsec:complexity-shgp}
\seqhierarchicalGP carries out a sequential training, one source at at time. In the following we analyse the complexity of these sequential steps individually.

\paragraph{Training of the first source}
The first source is a simple GP, with training complexity $\mathcal{O}(N_1^3)$ and prediction complexity $\mathcal{O}(N_1^2)$.

\paragraph{Training of the $n$th source}
In order to train the $n$th source, we need to evaluate the posterior mean and covariance of the $n-1$th source at the data points $X_n$. This requires evaluating the posterior means of all underlying sources at the same points, resulting in  $\mathcal{O}(N_{n} \sum_{\nu=1}^{n-1} N_{\nu})$ multiplications.

In addition we need to invert the covariance matrix $\left[\Sigma_{n}(X_{n}, X_{n}) + \sigma_{n}^2\mathbf{I}\right]$ which is of order $\mathcal{O}(N_{n}^3)$. In order to obtain the covariance matrix, we need to evaluate the posterior covariance of the $n-1$th source $X_n$
\begin{align}
\label{eq:post-cov-nm1}
\Sigma_{n}(X_n, X_n) 
    &= k_{n}(X_n, X_n) + \Sigma_{n-1}^{\text{post}}(X_n, X_n), \nonumber \\ 
\Sigma_{n-1}^{\text{post}}(X_n, X_n)
    &=\Sigma_{n-1}(X_n, X_n) \nonumber \\ 
    &-\Sigma_{n-1}(X_n, X_{n-1})
    \left[\Sigma_{n-1}(X_{n-1}, X_{n-1}) + \sigma_{n-1}^2\mathbf{I}\right]^{-1}
    \Sigma_{n-1}(X_{n-1}, X_n)
\end{align}
Inverting $\left[\Sigma_{n-1}(X_{n-1}, X_{n-1}) + \sigma_{n-1}^2\mathbf{I}\right]$ is part of the training of the $n-1$ source. 

From \cref{eq:post-cov-nm1} we can see that we need to evaluate $\Sigma_{n-i}^{\text{post}}(X_n, X_n)\, \forall i\in\{1, 2, ..., n-1\}$. This requires $\mathcal{O}(\sum_{i=1}^{n-1} (N_{n}^2 N_{n-i} + N_{n} N_{n-i}^2))$ multiplications and the calculation of $\Sigma_{n-i-1}^{\text{post}}(X_n, X_{n-i}) \, \forall i\in\{1, 2, ..., n-2\}$. The later has complexity $\mathcal{O}(\sum_{i=1}^{n-2} (N_{n} N_{n-i} N_{n-i-1} + \min(N_{n}, N_{n-i}) N_{n-i-1}^2))$ plus the cost of calculating $\Sigma_{n-i-j}^{\text{post}}(X_n, X_{n-i}) \, \forall j\in\{2, 3, ..., n-i-1\}$. Hence, we actually need the complexity of $\Sigma_{n-i-j}^{\text{post}}(X_n, X_{n-i}) \,\forall j\in\{1, 2, ..., n-i-1\}\, \forall i\in\{1, 2, ..., n-2\}$ which is given by $\mathcal{O}(\sum_{i=1}^{n-2} \sum_{j=1}^{n-i-1} (N_{n} N_{n-i} N_{n-i-j} + \min(N_{n}, N_{n-i}) N_{n-i-j}^2))$. All together we have a training complexity of
\begin{align}
\label{eq:post_cov_cost-general}
&\mathcal{O}\left(
    N_n^3
	+ \sum_{i=1}^{n-1} (N_{n}^2 N_{n-i} + N_{n} N_{n-i}^2)
	+ \sum_{i=1}^{n-2} \sum_{j=1}^{n-i-1} (N_{n} N_{n-i} N_{n-i-j} + \min(N_{n}, N_{n-i}) N_{n-i-j}^2)
	\right) \nonumber \\
&=\mathcal{O}\left(
	N_n^3
	+ N_{n}^2 \sum_{i=1}^{n-1} N_{i} 
	+ N_{n} \left(\sum_{i=1}^{n-1} N_{i}\right)^2
	\right).
\end{align}

\paragraph{Training of the target}
Training of the target is analogous to training the $n$th source. Setting $n=n_s+1$, we have complexity $\mathcal{O}(N_t^3 + N_t^2 N_s + N_t N_s^2)$.

\paragraph{Prediction from the target}
The prediction at a new query point $x_*$ is dominated by evaluating the covariance matrix. We can read-off this complexity from \cref{eq:post_cov_cost-general} by ignoring the $N_n^3$ term and setting $n=n_s+2$ and $N_{n_s+2}=1$, giving $\mathcal{O}((N_{t} + N_s)^2)$.


\subsection{\DEFstackGP (\stackGP)}
The joint kernel for $n_s$ sources and the target in \DEFstackGP is defined by \cref{eq:general_kernel-general} with
\begin{align}
    [\mathbf{W}_{\nu}]_{i,j} &= \delta_{i,\nu}\delta_{j,\nu}.
\end{align}
The number of hyperparameters is hence the number of hyperparameters in the kernel functions, which is of order $\mathcal{O}(n_s)$.

\paragraph{Training of the first source}
The first source is a simple GP, with training complexity $\mathcal{O}(N_1^3)$ and prediction complexity $\mathcal{O}(N_1^2)$.

\paragraph{Training of the $n$th source}
In order to train the $n$th source, we need to evaluate the posterior mean of the $n-1$th source at the data points of the $n$th source. This in turn requires the evaluating the posterior means of all underlying sources at the same points, all together  $\mathcal{O}(N_{n} \sum_{\nu=1}^{n-1} N_{\nu})$ multiplications. Except for this, the complexity is the same as in a simple GP with $N_n$ data points, hence all together we get a training complexity of $\mathcal{O}(N_{n}^3 + N_n \sum_{\nu=1}^{n-1} N_{\nu})$.

\paragraph{Training of the target}
Training of the target is analogous to training the $n$th source. Setting $n=n_s+1$, we have complexity $\mathcal{O}(N_t^3 + N_t N_s)$.

\paragraph{Prediction from the target}
In order to predict mean of the target model at a new query point $x_*$, we need to evaluate the posterior means of all sources at $x_*$, requiring $\mathcal{O}(N_s)$ multiplications. For predicting the covariance, the complexity is the same as for a simple GP $\mathcal{O}(N_{t}^2)$. The overall prediction complexity is hence $\mathcal{O}(N_{t}^2 + N_s)$.

\subsection{\DEFboostedGP (\boostedGP)}
\label{ap:subsec:complexity-bhgp}
To the best of our knowledge in \DEFboostedGP there is no unique generalization to  $n_s>1$ sources, because of the heuristic aggregation of uncertainty. Below we outline several avenues that one could consider, which differ in terms of complexity. The most straightforward approach is to do everything as in \DEFstackGP, and add a boosting term to the posterior prediction of the target task. The kernel for that is simply given by
\begin{equation}
\label{eq:boosted_kernel_multisource}
    k((\x, i), (\x', j)) = \sum_{\nu=1}^{n_s+2} [\mathbf{W}_\nu]_{i, j} k_\nu(x, x') + 
    \delta_{xx'}\delta_{ij}\sigma_i^2 \qquad \forall i, j \in \{1, 2, ..., n_s +2\}.
\end{equation}
In this single layer boosting, we have $[\mathbf{W}_\nu]_{ij} = \delta_{i\nu} \delta_{j\nu}$, $k_{n_s+2} = k_{n_s+1} + \Sigma_{n_s+1}^{\spl}$, and $\sigma_{n_s+2} = 0$. Here $\nu, i, j=n_s+2$ describes the query points and 
\begin{align}
\label{eq:boosting_term_appendix}
\Sigma_{n_s+1}^{\spl}(\x_*, \x_*)
    &= 
    \Sigma_{n_s}^{\text{post}}(\x_*, \x_*)
    + \alpha_{n_s} \Sigma_{n_s}^{\text{post}}(\X_{n_s+1}, \X_{n_s+1}) \alpha_{n_s}^T \nonumber \\
    &\quad -\alpha_{n_s} \Sigma_{n_s}^{\text{post}}(\X_{n_s+1}, \x_*)
    -\Sigma_{n_s}^{\text{post}}(\x_*, \X_{n_s+1}) \alpha_{n_s}^T, \\
\alpha_{n_s}
    &= k_{n_s+1}(\x_*, \X_{n_s+1})\left(k_{n_s+1}(\X_{n_s+1}, \X_{n_s+1}) + \sigma_{n_s+1}^2\mathbb 1\right)^{-1}.
\end{align}
In that case, the boosting is only done in the last layer. The complexity of \DEFboostedGP has two contributions. One, the complexity of \DEFstackGP and the complexity of calculating the boosting term. For the boosting term, we need to calculate $\Sigma_{n_s}^{\text{post}}(\X_{n_s+1}, \X_{n_s+1})$ which is of order $\mathcal{O}\left( N_{n_s+1}^2 N_{n_s} + N_{n_s+1} N_{n_s}^2 \right)$, this can be done during training, since it does not depend on the query point. During prediction we need to evaluate $\alpha_{n_s}$ which is of order $\mathcal{O}\left(N_{n_s+1}^2\right)$ because the matrix inversion can be done as part of the training. We also need to to evaluate $\Sigma_{n_s}^{\text{post}}(\x_*, \x_*)$ and $\Sigma_{n_s}^{\text{post}}(\x_*, \X_{n_s+1})$ which is of order $\mathcal{O}\left(N_{n_s+1} N_{n_s} + N_{n_s}^2\right)$. To conclude the calculation of the boosting term, we need to evaluate \cref{eq:boosting_term_appendix} which is of order $\mathcal{O}\left(N_{n_s+1}^2\right)$. All together we have training complexity $\mathcal{O}\left(\sum_{\nu=1}^{n_s+1} N_\nu^3 + N_{n_s+1}^2 N_{n_s} + N_{n_s+1} N_{n_s}^2\right)$ and prediction complexity $\mathcal{O}\left(N_{n_s+1}^2 + N_{n_s+1} N_{n_s} + N_{n_s}^2\right) + \sum_{\nu=1}^{n_s} N_\nu)$.

An alternative approach which we follow in our experiments is to add such boosting terms at each intermediate prediction in the hierarchy by substituting $\Sigma_{n_s}^{\text{post}} \rightarrow \Sigma_{n_s}^{\text{post}} + \Sigma_{n_s}^{\spl}$ recursively in \cref{eq:boosting_term_appendix}. Note that $\Sigma_{1}^{\text{post}}=0$.
Note that the likelihood of the boosted model is the same as of \DEFstackGP. Thus, this change only affects the prediction. However, before being able to make predictions, we now have to pre-calculate additional terms. In particular, this requires to recursively evaluate the posterior covariances, very similarly as in \cref{eq:post-cov-nm1}. This leads to an additional training complexity of $\mathcal{O}(N_t^2 N_s + N_t N_s^2)$. During prediction, the additional cost comes from the boost term which involves recursive evaluation of the posterior covariance terms containing the query point. This increases the total prediction cost to  $\mathcal{O}((N_t +N_s)^2)$.

\begin{table}
  \caption{Computational complexity of the algorithms. Meta training is performed once before starting the optimization. Training the target model is carried out once at every optimization step. Prediction occurs multiple times during each BO step when optimizing the acquisition function. The kernel-based techniques model the data from all tasks jointly and do therefore not have a meta-training stage.
  }
  \label{tab:computational_complexity_multiple_sources}
  \centering
  \begin{tabular}{lllll}
    \toprule
    Models & Stage & Complexity \\
    \midrule
    \multirow{3}{4em}{\multitaskGP} & meta training & N/A \\
    & target training & $\mathcal{O}\left[ (N_t+N_s)^3 \right]$ \\
    & prediction & $\mathcal{O}\left[ (N_t+N_s)^2 \right]$\\
    \midrule
    \multirow{3}{4em}{\multitasksinglekGP} & meta training & N/A \\
    & target training & $\mathcal{O}\left[ (N_t+N_s)^3 \right]$ \\
    & prediction & $\mathcal{O}\left[ (N_t+N_s)^2 \right]$\\
    \midrule
    \multirow{3}{4em}{\weightedSourceGP} & meta training & N/A \\
    & target training & $\mathcal{O}\left[ N_t^3 + N_t^2 N_s + N_t N_s^2 + \sum_{\nu=1}^{n_s} N_\nu^3 \right]$ \\
    & prediction & $\mathcal{O}\left[ (N_t+N_s)^2 \right]$\\
    \midrule
    \multirow{3}{4em}{\hierarchicalGP} & meta training & N/A \\
    & target training & $\mathcal{O}\left[ (N_t+N_s)^3 \right]$ \\
    & prediction & $\mathcal{O}\left[ (N_t+N_s)^2 \right]$\\
    \midrule
    \multirow{3}{4em}{\seqhierarchicalGP} & meta training & $\mathcal{O}\left[ N_s^3 \right]$ \\
    & target training & $\mathcal{O}\left[ N_t^3 + N_t^2 N_s + N_t N_s^2 \right]$ \\
    & prediction & $\mathcal{O}\left[ (N_t+N_s)^2 \right]$\\
    \midrule
    \multirow{3}{4em}{\stackGP} & meta training & $\mathcal{O}\left[ \sum_{\nu=1}^{n_s} N_\nu^3 + N_s^2\right]$ \\
    & target training & $\mathcal{O}\left[ N_t^3 + N_t N_s\right]$ \\
    & prediction & $\mathcal{O}\left[ N_t^2 + N_s \right]$\\
    \midrule
    \multirow{3}{4em}{\boostedGP} & meta training & $\mathcal{O}\left[ N_s^3 \right]$ \\
    & target training & $\mathcal{O}\left[ N_t^3 + N_t^2 N_s + N_t N_s^2 \right]$ \\
    & prediction & $\mathcal{O}\left[ (N_t+N_s)^2 \right]$\\
    \bottomrule
  \end{tabular}
\end{table}

\newpage
\section{DETAILS ON EXPERIMENTS}
\label{ap:details_experiments}
All the experiments use default BO parameters with Upper Confidence Bound (UCB) and exploration coefficient $\beta=3$ as acquisition function. The source and target functions are sampled randomly from the function family, and the source data are sampled randomly from each source task, see \cref{ap:synthetic_functions} for details. BO is used to optimize the target function from scratch, i.e., without initial samples. A certain amount of i.i.d. zero-mean Gaussian observational noise is added during the data generation process with a standard deviation specified in each figure caption. 

All GP models are based on the squared-exponential kernel with automatic relevance determination. The GP hyperparameters are optimized by maximizing the likelihood function for the observed data using the L-BFGS-B optimizer with 10 initial guesses, $\mathbf{x}_0$, where each instance is sampled from
\begin{equation}
x_0 = \log\left(1 + \exp\left(x_0'\right)\right),\qquad x_0' \sim \mathcal{N}(0, 1).
\end{equation}
This generic optimization strategy is possible since we normalize the observed data to zero-mean, unit-variance. All other details regarding training and prediction are kept fixed to GPy's defaults \citep{gpy2014}. The only exception is the OpenML SVM benchmark, where, for complexity reasons, we only use 3 initial guesses and a maximum of 20 gradient-ascent steps for the optimization of the likelihood function of \hierarchicalGP and \weightedSourceGP.

\subsection{Details on the Synthetic Function Families}
\label{ap:synthetic_functions}
\paragraph{The Forrester Family} 
The Forrester function  is a one-dimensional synthetic function with one global minimum, one local minimum, and a zero-gradient inflection point. The function is given by
\begin{equation}
f(x; a, b, c) = a\left(6x-2\right)^2\sin\left(12x-4\right)+b\left(x-\frac{1}{2}\right)-c,\quad x\in[0, 1].
\end{equation}
The original function is given by $a=1, b=0, c=0$. In this paper, a family of functions is formed by choosing the following probability distributions for the parameters $(a, b, c)$:
\begin{equation}
    a\sim\mathcal{U}(0.2, 3),\quad b\sim\mathcal{U}(-5, 15), \quad c\sim\mathcal{U}(-5, 5),
\label{eq:forrester_family}
\end{equation}
where $\mathcal{U}(\cdot,\cdot)$ denotes the uniform distribution. The Forrester family is therefore defined over a three-dimensional uniform distribution.

For generating the data of $n_s$ source data sets, we draw $n_s$ random tasks using \cref{eq:forrester_family}, and sample a given number of points per task using a random distribution. In \cref{fig:experiments_single_source}, 20 points per task are sampled.

\paragraph{The Alpine Family}
The Alpine function  is a one-dimensional synthetic function with one global minimum and three local minima. The function is given by
\begin{equation}
f(x; s) = x\sin(x+\pi+s)+0.1x,\quad x\in[-10, 10].
\end{equation}
The original function is given by $s=0$. In this paper, the Alpine function is used for the multi-source experiments in \cref{fig:experiments_multi_source}. There, $s=0$ is used for the target function, while the five source functions are given by $s=k\pi/12, k=1,...,5$. Twenty random points are sampled for each source function. Note that this benchmark is identical to the one presented in \citep[Sec.~5.1]{feurer2018rgpe}.

In the single-source experiments presented in \cref{fig:experiments_single_source}, the target function is also set to $s=0$, while the source function is sampled randomly from the five choices above.

\paragraph{The Branin Family}
The Branin function is a two-dimensional synthetic function defined as
\begin{equation}
f(x_1, x_2; a, b, c, r, s, t) = a(x_2-bx_1^2+cx_1-r)+s(1-t)\cos(x_1)+s,\quad x_1\in\left[-5, 10\right], x_2\in\left[0, 15\right]
\label{eq:branin_fun}
\end{equation}
A family of functions is formed by choosing the following probability distributions for the parameters $(a, b, c, r, s, t)$:
\begin{equation}
a\sim\mathcal{U}(0.5, 1.5), b\sim\mathcal{U}(0.1, 0.15), c\sim\mathcal{U}(1, 2), r\sim\mathcal{U}(5, 7), s\sim\mathcal{U}(8, 12),
t\sim\mathcal{U}(0.03, 0.05).
\label{eq:branin_family}
\end{equation}
The Branin family is therefore defined over a six-dimensional uniform distribution. For generating the data of $n_s$ source data sets, we draw $n_s$ random tasks using \cref{eq:branin_family}, and sample a given number of points per task using a random distribution. In \cref{fig:experiments_single_source}, 40 points per task are sampled.

\paragraph{The Hartmann3 Family}
The Hartmann3 function is a sum of four three-dimensional Gaussian distributions and is defined by
\begin{equation}
f(\mathbf{x}; \bm{\alpha}) = -\sum_{i=1}^4\alpha_i\exp\left(-\sum_{j=1}^3 A_{i,j}\left(x_j-P_{i,j}\right)^2 \right),\quad \mathbf{x} \in \left[0, 1\right]^3,
\label{eq:hartmann3_fun}
\end{equation}
with
\begin{equation*}
\mathbf{A} = \begin{bmatrix}
3.0 &10  &30 \\ 
0.1 &10  &35 \\ 
3.0 &10  &30 \\ 
0.1 &10  &35 
\end{bmatrix},\quad \mathbf{P} = 10^{-4}
\begin{bmatrix}
3689 &1170  &2673 \\ 
4699 &4387  &7470 \\ 
1091 &8732  &5547 \\ 
381  &5743  &8828 
\end{bmatrix}.
\end{equation*}
The original Hartmann3 function is given by $\bm{\alpha}=(1.0, 1.2, 3.0, 3.2)^T$. In this paper, a family of functions is formed by choosing the following probability distributions for the parameters $\bm{\alpha}=(\alpha_1, \alpha_2, \alpha_3, \alpha_4)^T$:
\begin{equation}
\alpha_1\sim\mathcal{U}(1.00, 1.02), \alpha_2\sim\mathcal{U}(1.18, 1.20),
\alpha_3\sim\mathcal{U}(2.8, 3.0),
\alpha_4\sim\mathcal{U}(3.2, 3.4).
\label{eq:hartmann3_family}
\end{equation}
The Hartmann3 family is therefore defined over a four-dimensional uniform distribution. For generating the data of $n_s$ source data sets, we draw $n_s$ random tasks using \cref{eq:hartmann3_family}, and sample a given number of points per task using a random distribution. In \cref{fig:experiments_single_source}, 60 points per task are sampled.

\paragraph{The Hartmann6 Family}
The Hartmann6 family is a sum of four six-dimensional Gaussian distributions and is defined by
\begin{equation}
f(\mathbf{x}; \bm{\alpha}) = -\sum_{i=1}^4\alpha_i\exp\left(-\sum_{j=1}^6 A_{i,j}\left(x_j-P_{i,j}\right)^2 \right),\quad \mathbf{x} \in \left[0, 1\right]^6,
\label{eq:hartmann6_fun}
\end{equation}
with
\begin{equation*}
\begin{split}
\mathbf{A} &= 
\begin{bmatrix}
10   & 3   & 17   & 3.5 & 1.7 & 8 \\
0.05 & 10  & 17   & 0.1 & 8   & 14 \\
3    & 3.5 & 1.7  & 10  & 17  & 8 \\
17   & 8   & 0.05 & 10  & 0.1 & 14
\end{bmatrix},\\
\mathbf{P} &= 10^{-4}
\begin{bmatrix}
1312 & 1696 & 5569 & 124  & 8283 & 5886 \\
2329 & 4135 & 8307 & 3736 & 1004 & 9991 \\
2348 & 1451 & 3522 & 2883 & 3047 & 6650 \\
4047 & 8828 & 8732 & 5743 & 1091 & 381
\end{bmatrix}.
\end{split}
\end{equation*}
The original Hartmann6 function is given by $\bm{\alpha}=(1.0, 1.2, 3.0, 3.2)^T$. In this paper, a family of functions is formed by choosing the following probability distributions for the parameters $\bm{\alpha}=(\alpha_1, \alpha_2, \alpha_3, \alpha_4)^T$:
\begin{equation}
\alpha_1\sim\mathcal{U}(1.00, 1.02), \alpha_2\sim\mathcal{U}(1.18, 1.20),
\alpha_3\sim\mathcal{U}(2.8, 3.0),
\alpha_4\sim\mathcal{U}(3.2, 3.4).
\label{eq:hartmann6_family}
\end{equation}
The Hartmann6 family is therefore defined over a four-dimensional uniform distribution. For generating the data of $n_s$ source data sets, we draw $n_s$ random tasks using \cref{eq:hartmann6_family}, and sample a given number of points per task using a random distribution. In \cref{fig:experiments_single_source} and \cref{fig:experiments_multi_source}, 120 and 60 points per task are sampled, respectively.

\subsection{Details on the Surrogate Meta-Learning Benchmarks}
\label{app:meta-learning-benchmarks}
The SVM benchmark tunes four hyperparameters: cost (float), gamma (float), degree (integer), and kernel type (discrete). The same data set was used in \citet{perrone2018scalable} for transfer-learning benchmarks. Further details are provided in \cref{tbl:surrogate_benchmark_datasets} and in \citet{kuhn2018openml}.

\begin{table}[h!]
\centering
\begin{tabular}{c c c c c c c c}
\hline
Scenario Name & \parbox{1cm}{OpenML flow id} & OpenML data set ids & $\min N_t $ & $\max N_t$ & $\sum_t N_t$ \\ [0.5ex]
\hline
OpenML\_SVM & 5891 & \parbox{3.2cm}{\center 10101, 145878, 146064, 14951, 34537, 3485, 3492, 3493, 3494, 37, 3889, 3891, 3899, 3902, 3903, 3913, 3918, 3950, 9889, 9914, 9946, 9952, 9967, 9971, 9976, 9978, 9980, 9983} & 987 & 3217 & 50217 \\[1.5cm]
%
\hline \\
\end{tabular}
\caption{
OpenML Random Bot surrogate benchmark test data set characteristics, selected from the full repertoire of $37$ data sets, cf. \citet[Table 2]{kuhn2018openml}.
$\min N_t$ ($\max N_t$) denotes the minimum (maximum) number of data points in a task, while $\sum_t N_t$ is the total number of points in the data set.
}
\label{tbl:surrogate_benchmark_datasets}
\end{table}

\newpage
\section{FURTHER EXPERIMENTS}
\subsection{Results on one-dimensional benchmark functions}
\label{ap:results_1d_benchmarks}
\begin{figure}
  \centering
  \includegraphics[width=\textwidth]{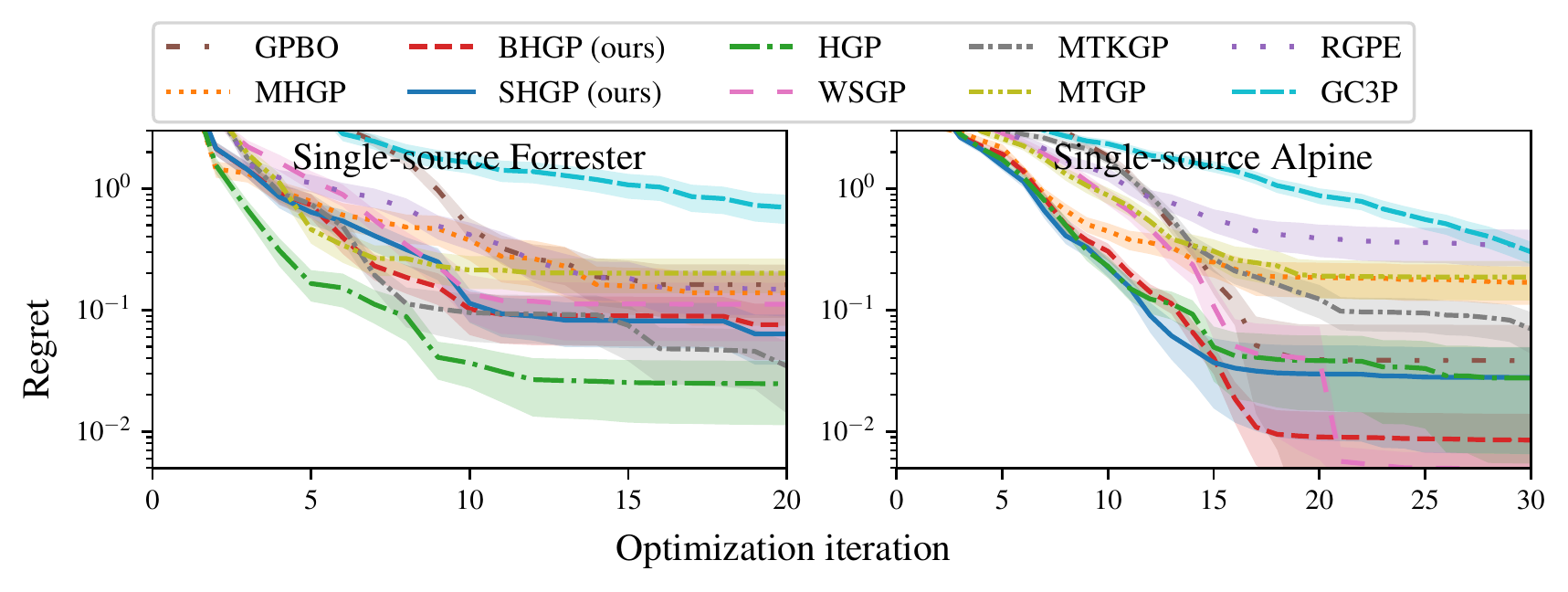}
  \caption{Performance on the single-source one-dimensional benchmarks: the Forrester (left) and Alpine (right) function families. The mean and standard error of the mean of the regret function are plotted. Due to enhanced stochasticity, the statistics are computed over 200 independent runs. The source data are sampled randomly from the source function and contain 20 points. I.i.d. observational noise of standard deviation $\sigma_s=\sigma_t=0.1$ is added during data generation.}
  \label{fig:single-source-one-dim-benchmarks}
\end{figure}

Results on the single-source one-dimensional Forrester and Alpine function families are presented in \cref{fig:single-source-one-dim-benchmarks}. The function families are described in \cref{ap:synthetic_functions}. These results are consistent with the discussion and analysis from \cref{sec:single_source_experiments}.

\subsection{Runtime Analysis}
\label{ap:runtime_analysis}
\begin{figure}
  \centering
  \includegraphics[width=0.7\textwidth]{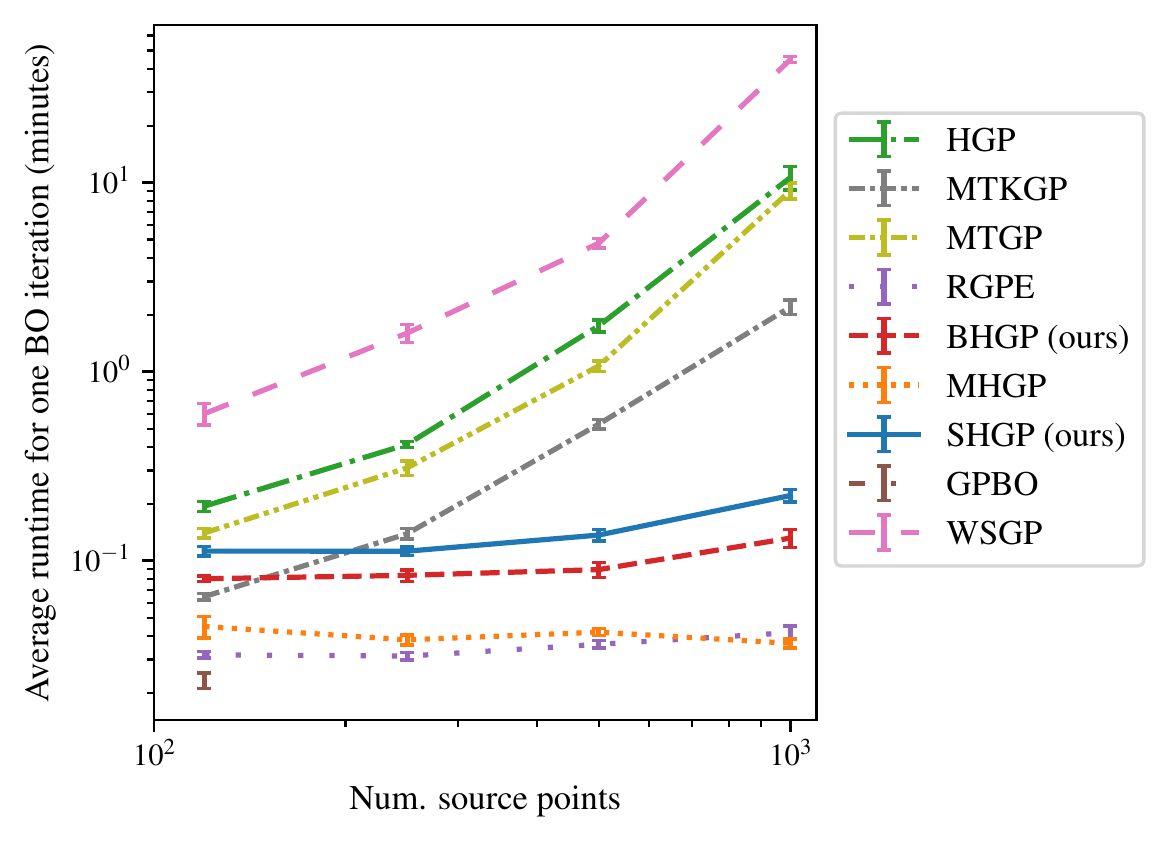}
  \caption{Average runtime per BO step versus the number of points in the source data set when optimizing on the Hartmann6 benchmark. The average is taken over the first twenty-five BO steps. The source data are sampled randomly from the source function. I.i.d. observational noise of standard deviation $\sigma_s = \sigma_t = 0.1$ is added during data generation. The mean and standard error of the mean are computed over 7 independent runs. The performance of GPBO is independent of the number of historical points.}
  \label{fig:runtime_entire_bo}
\end{figure}

Here we present runtime studies for the entire optimization procedure, which includes (i) training the models including optimizing their hyperparameters, and (ii) optimization of the acquisition function. We time the runs by optimizing the Hartmann6 benchmark function. We consider one single source task with the same configuration as in \cref{fig:experiments_single_source}. Results are shown in \cref{fig:runtime_entire_bo} and are consistent with training runtimes reported in \cref{fig:experiments_num_source_and_noise}. This reinforces that model training is more expensive than acquisition-function optimization since the former has steeper complexity than the latter for the Bayesian models, see \cref{tab:computational_complexity,tab:computational_complexity_multiple_sources}.

\subsection{Impact of Observational Noise and Propagation of Uncertainty: Ablation Studies}
\label{ap:ablation_studies}
We perform a systematic experimental study on the impact of observational noise and amount of source data on the algorithm performance. The three regimes discussed in \cref{subsec:exp_impact_noise} can be clearly observed for the Hartmann3 function family in \cref{fig:experiments_num_source_and_noise}.

\begin{figure}
  \centering
  \includegraphics[width=\textwidth]{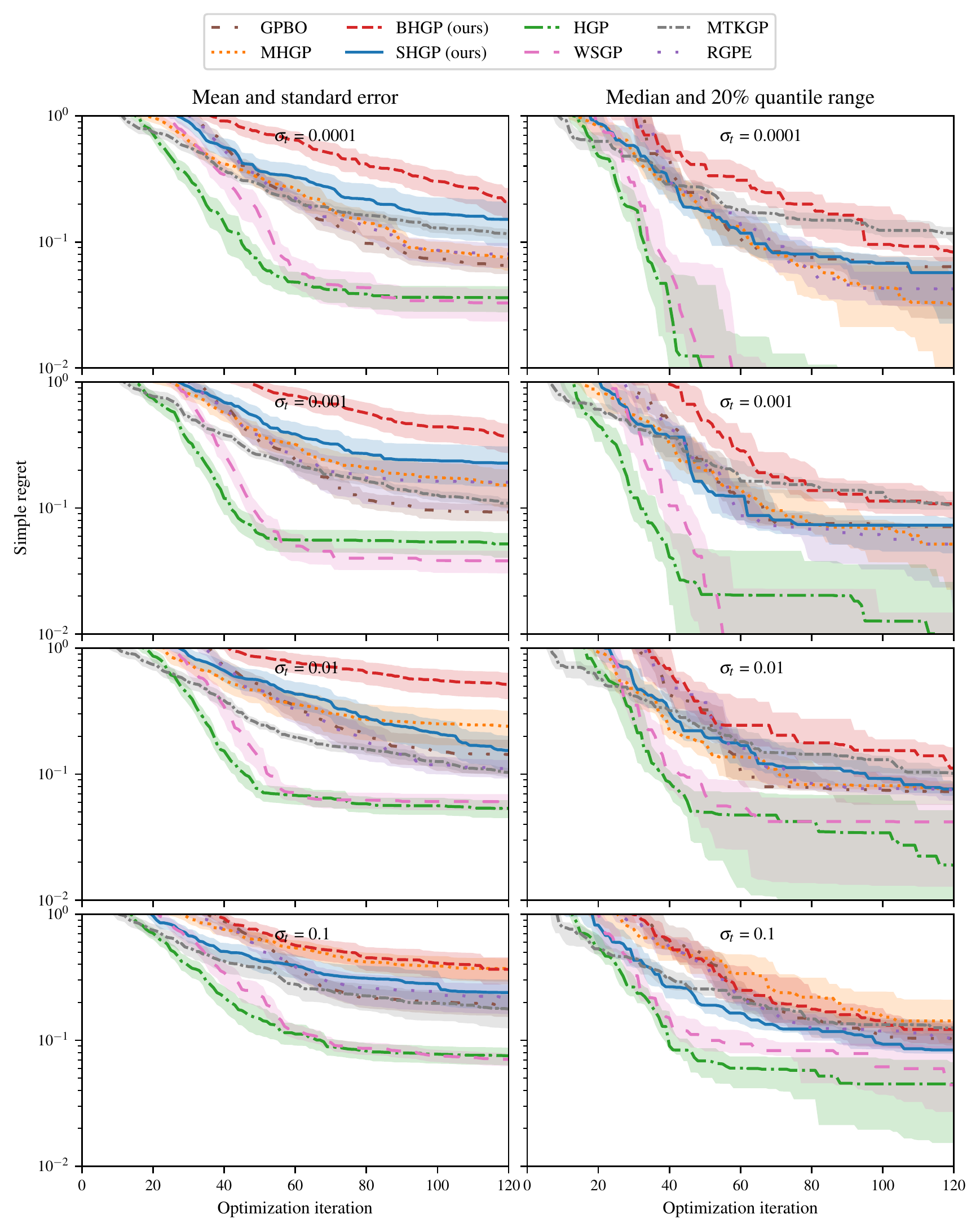}
  \caption{Performance on the single-source Hartmann6 function family for 30 points in the source data set. We vary the amount of observational noise, $\sigma_s=\sigma_t$, from $0.001$ to $0.1$. The left panel depicts the mean $\pm$ standard error of the mean. The right panel shows the median and 20\% quantile range.}
  \label{fig:sup:hartmann6_30}
\end{figure}

\begin{figure}
  \centering
  \includegraphics[width=\textwidth]{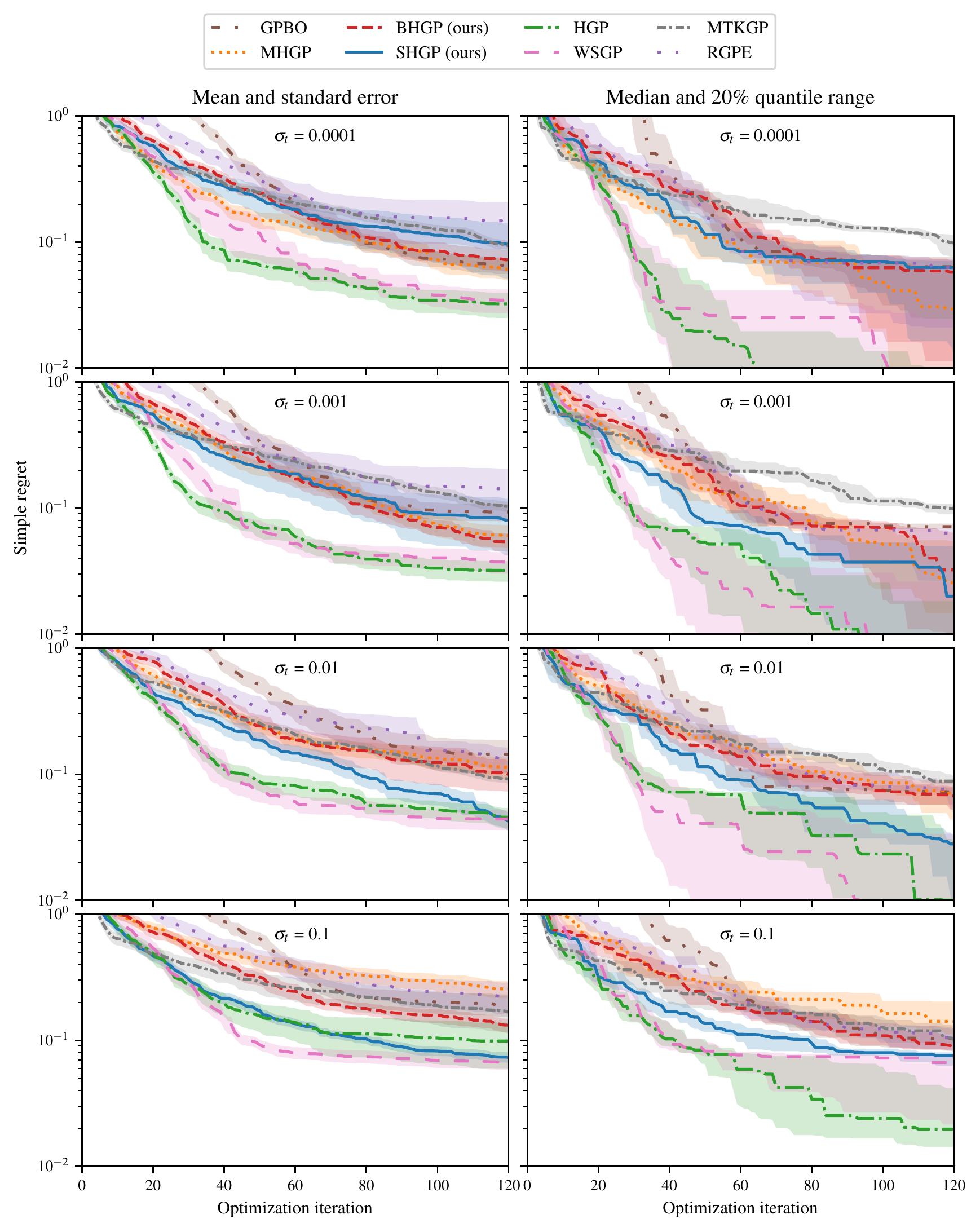}
  \caption{Performance on the single-source Hartmann6 function family for 60 points in the source data set. We vary the amount of observational noise, $\sigma_s=\sigma_t$, from $0.001$ to $0.1$. The left panel depicts the mean $\pm$ standard error of the mean. The right panel shows the median and 20\% quantile range.}
  \label{fig:sup:hartmann6_60}
\end{figure}

\begin{figure}
  \centering
  \includegraphics[width=\textwidth]{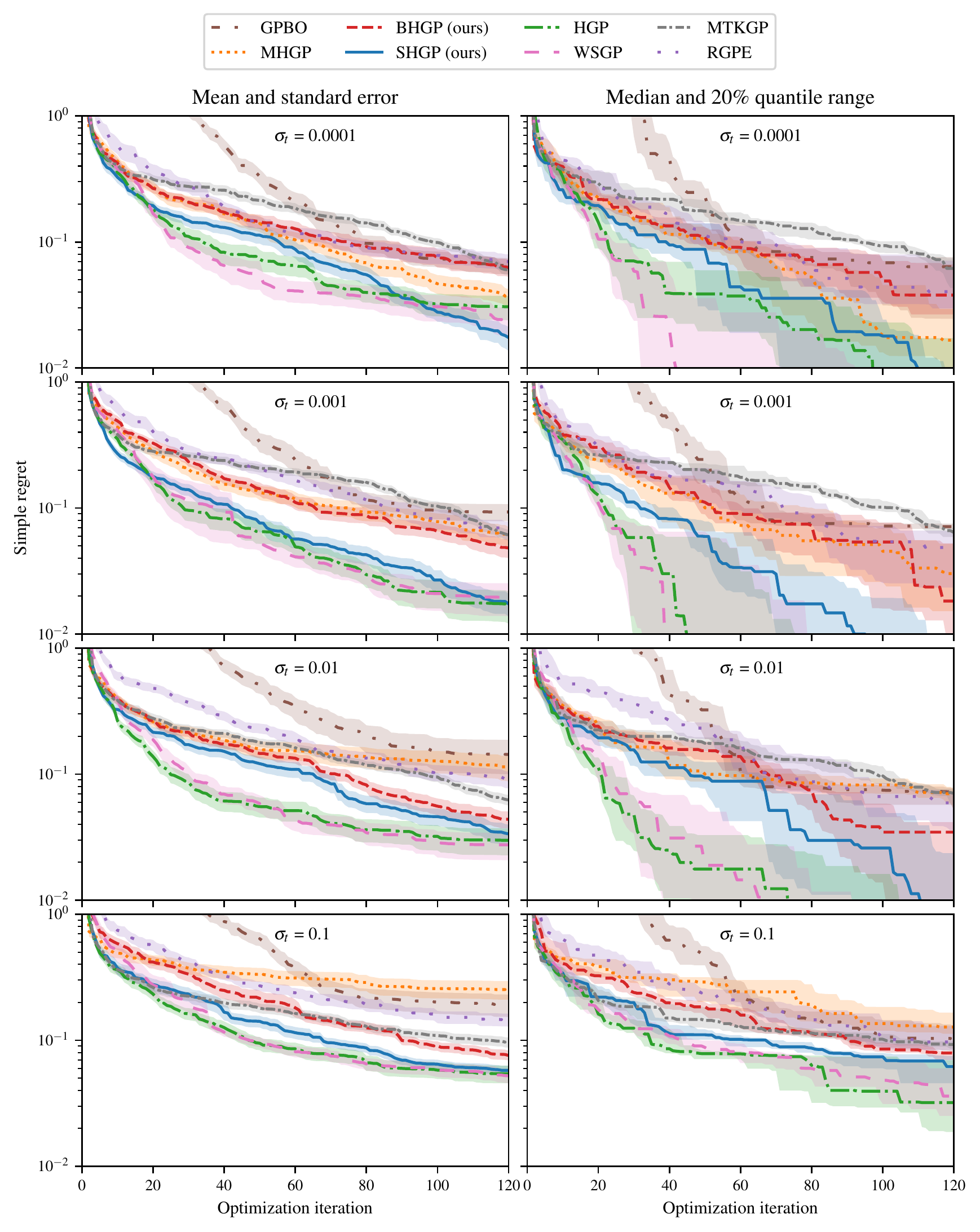}
  \caption{Performance on the single-source Hartmann6 function family for 120 points in the source data set. We vary the amount of observational noise, $\sigma_s=\sigma_t$, from $0.001$ to $0.1$. The left panel depicts the mean $\pm$ standard error of the mean. The right panel shows the median and 20\% quantile range.}
  \label{fig:sup:hartmann6_120}
\end{figure}

\begin{figure}
  \centering
  \includegraphics[width=\textwidth]{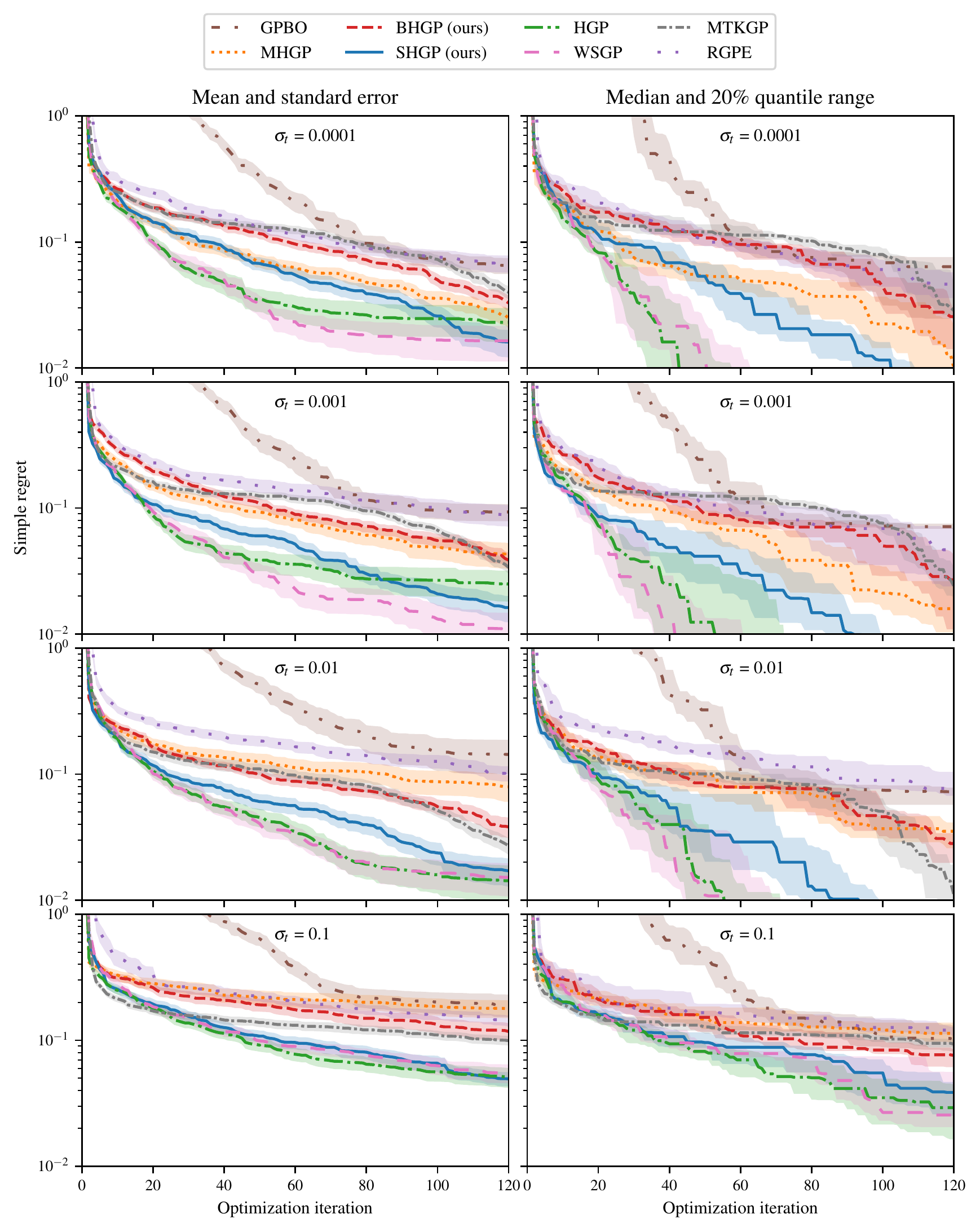}
  \caption{Performance on the single-source Hartmann6 function family for 250 points in the source data set. We vary the amount of observational noise, $\sigma_s=\sigma_t$, from $0.001$ to $0.1$. The left panel depicts the mean $\pm$ standard error of the mean. The right panel shows the median and 20\% quantile range.}
  \label{fig:sup:hartmann6_250}
\end{figure}

\begin{figure}
  \centering
  \includegraphics[width=\textwidth]{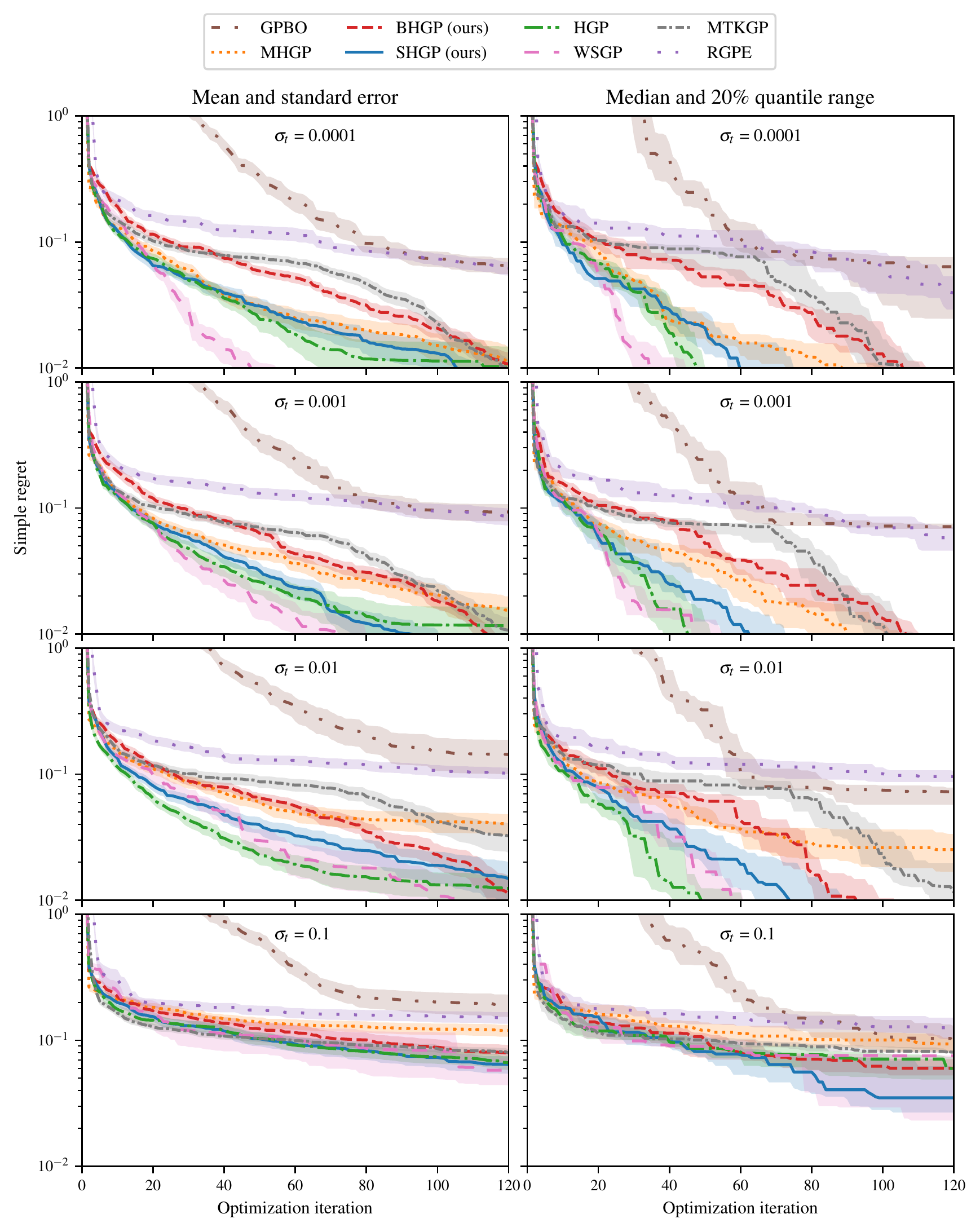}
  \caption{Performance on the single-source Hartmann6 function family for 500 points in the source data set. We vary the amount of observational noise, $\sigma_s=\sigma_t$, from $0.001$ to $0.1$. The left panel depicts the mean $\pm$ standard error of the mean. The right panel shows the median and 20\% quantile range.}
  \label{fig:sup:hartmann6_500}
\end{figure}

Here, we showcase the dynamics of these three regimes for the challenging Hartmann6 function family in \cref{fig:sup:hartmann6_30,fig:sup:hartmann6_60,fig:sup:hartmann6_120,fig:sup:hartmann6_250,fig:sup:hartmann6_500}. The first regime, in which the amount of source data is insufficient to build a model that faithfully describes the global shape of the source function, is presented in \cref{fig:sup:hartmann6_30}, where only 30 data points are considered. Here, algorithms that model the data jointly (\hierarchicalGP, \weightedSourceGP) have a clear advantage.  A crossover to the intermediate regime, in which the source data describe the source function much better but with significant uncertainty, happens gradually in \cref{fig:sup:hartmann6_60,fig:sup:hartmann6_120,fig:sup:hartmann6_250}, where the source data are increased from 60 to 250 points. Here, our relatively lightweight \seqhierarchicalGP is competitive and on par in terms of performance with the more general \hierarchicalGP. The third regime is reached in \cref{fig:sup:hartmann6_500}, where the performance of all methods except GPBO and RGPE converge to similar performance.

The observational noise adds a further level of complexity to these dynamics. In essence, the observational noise controls the boundary between the three regimes: the more noise is present, the more data points are required to describe the function for a fixed model quality. In other words, observational noise leads to enhanced model uncertainty, whose propagation is key to obtaining a high-quality model. It is therefore expected that (i) the performance gap between Bayesian and non-Bayesian techniques becomes ever bigger for increased observational noise, and (ii) our developed techniques, \seqhierarchicalGP and \boostedGP, particularly excel for elevated observational noise compared to algorithms that do not propagate uncertainty like \stackGP and RGPE. These aspects can be observed in \cref{fig:sup:hartmann6_60,fig:sup:hartmann6_120}.

\newpage
\section{IMPLEMENTATION AND COMPUTATIONAL RESOURCES}
\label{ap:resources}
All experiments were run on a HPC cluster, where each individual experiment used a single Intel Xeon CPU. All experiments (including early debugging and evaluations) amounted to a total of 154353 hours, which corresponds to roughly 17.6 years if the jobs ran sequentially. Most of this compute was required to ensure reproducibility (multiple random seeds per job and ablation studies over the effects of parameters). The cluster is part of a carbon-neutral infrastructure and does not leave a carbon footprint.

\end{document}